\documentclass{article}
\PassOptionsToPackage{colorlinks=true, allcolors=blue}{hyperref}

\usepackage[preprint]{neurips_2026}

\usepackage[no-math]{fontspec}
\usepackage{hyperref}
\usepackage{xurl}
\hypersetup{
  pdftitle={Formal Conjectures: An Open and Evolving Benchmark for Verified Discovery in Mathematics},
  pdfauthor={Moritz Firsching, Paul Lezeau, Salvatore Mercuri, Miklós Z. Horváth, Yaël Dillies, Calle Sönne, Eric Wieser, Fred Zhang, Thomas Hubert, Blaise Agüera y Arcas, Pushmeet Kohli}
}
\usepackage{booktabs}
\usepackage{amsfonts}
\usepackage{amsmath}
\usepackage{amssymb}
\usepackage{nicefrac}
\usepackage{microtype}
\usepackage[svgnames]{xcolor}
\usepackage{graphicx}
\usepackage{listings}
\usepackage{multirow}
\usepackage{float}
\usepackage{fvextra}
\usepackage{csquotes}

\usepackage{tikz}
\usetikzlibrary{shapes.geometric, arrows.meta, positioning, fit, backgrounds, calc}
\usepackage{pgfplots}
\pgfplotsset{compat=1.18}

\usepackage{wrapfig}

\usepackage[disable]{todonotes}
\usepackage{comment}
\setuptodonotes{color=blue!30, fancyline}

\makeatletter
\def\PYG@reset{\let\PYG@it=\relax \let\PYG@bf=\relax%
    \let\PYG@ul=\relax \let\PYG@tc=\relax%
    \let\PYG@bc=\relax \let\PYG@ff=\relax}
\def\PYG@tok#1{\csname PYG@tok@#1\endcsname}
\def\PYG@toks#1+{\ifx\relax#1\empty\else%
    \PYG@tok{#1}\expandafter\PYG@toks\fi}
\def\PYG@do#1{\PYG@bc{\PYG@tc{\PYG@ul{%
    \PYG@it{\PYG@bf{\PYG@ff{#1}}}}}}}
\def\PYG#1#2{\PYG@reset\PYG@toks#1+\relax+\PYG@do{#2}}

\@namedef{PYG@tok@w}{\def\PYG@tc##1{\textcolor[rgb]{0.73,0.73,0.73}{##1}}}
\@namedef{PYG@tok@c}{\let\PYG@it=\textit\def\PYG@tc##1{\textcolor[rgb]{0.24,0.48,0.48}{##1}}}
\@namedef{PYG@tok@cp}{\def\PYG@tc##1{\textcolor[rgb]{0.61,0.40,0.00}{##1}}}
\@namedef{PYG@tok@k}{\let\PYG@bf=\textbf\def\PYG@tc##1{\textcolor[rgb]{0.00,0.50,0.00}{##1}}}
\@namedef{PYG@tok@kp}{\def\PYG@tc##1{\textcolor[rgb]{0.00,0.50,0.00}{##1}}}
\@namedef{PYG@tok@kt}{\def\PYG@tc##1{\textcolor[rgb]{0.69,0.00,0.25}{##1}}}
\@namedef{PYG@tok@o}{\def\PYG@tc##1{\textcolor[rgb]{0.40,0.40,0.40}{##1}}}
\@namedef{PYG@tok@ow}{\let\PYG@bf=\textbf\def\PYG@tc##1{\textcolor[rgb]{0.67,0.13,1.00}{##1}}}
\@namedef{PYG@tok@nb}{\def\PYG@tc##1{\textcolor[rgb]{0.00,0.50,0.00}{##1}}}
\@namedef{PYG@tok@nf}{\def\PYG@tc##1{\textcolor[rgb]{0.00,0.00,1.00}{##1}}}
\@namedef{PYG@tok@nc}{\let\PYG@bf=\textbf\def\PYG@tc##1{\textcolor[rgb]{0.00,0.00,1.00}{##1}}}
\@namedef{PYG@tok@nn}{\let\PYG@bf=\textbf\def\PYG@tc##1{\textcolor[rgb]{0.00,0.00,1.00}{##1}}}
\@namedef{PYG@tok@ne}{\let\PYG@bf=\textbf\def\PYG@tc##1{\textcolor[rgb]{0.80,0.25,0.22}{##1}}}
\@namedef{PYG@tok@nv}{\def\PYG@tc##1{\textcolor[rgb]{0.10,0.09,0.49}{##1}}}
\@namedef{PYG@tok@no}{\def\PYG@tc##1{\textcolor[rgb]{0.53,0.00,0.00}{##1}}}
\@namedef{PYG@tok@nl}{\def\PYG@tc##1{\textcolor[rgb]{0.46,0.46,0.00}{##1}}}
\@namedef{PYG@tok@ni}{\let\PYG@bf=\textbf\def\PYG@tc##1{\textcolor[rgb]{0.44,0.44,0.44}{##1}}}
\@namedef{PYG@tok@na}{\def\PYG@tc##1{\textcolor[rgb]{0.41,0.47,0.13}{##1}}}
\@namedef{PYG@tok@nt}{\let\PYG@bf=\textbf\def\PYG@tc##1{\textcolor[rgb]{0.00,0.50,0.00}{##1}}}
\@namedef{PYG@tok@nd}{\def\PYG@tc##1{\textcolor[rgb]{0.67,0.13,1.00}{##1}}}
\@namedef{PYG@tok@s}{\def\PYG@tc##1{\textcolor[rgb]{0.73,0.13,0.13}{##1}}}
\@namedef{PYG@tok@sd}{\let\PYG@it=\textit\def\PYG@tc##1{\textcolor[rgb]{0.73,0.13,0.13}{##1}}}
\@namedef{PYG@tok@si}{\let\PYG@bf=\textbf\def\PYG@tc##1{\textcolor[rgb]{0.64,0.35,0.47}{##1}}}
\@namedef{PYG@tok@se}{\let\PYG@bf=\textbf\def\PYG@tc##1{\textcolor[rgb]{0.67,0.36,0.12}{##1}}}
\@namedef{PYG@tok@sr}{\def\PYG@tc##1{\textcolor[rgb]{0.64,0.35,0.47}{##1}}}
\@namedef{PYG@tok@ss}{\def\PYG@tc##1{\textcolor[rgb]{0.10,0.09,0.49}{##1}}}
\@namedef{PYG@tok@sx}{\def\PYG@tc##1{\textcolor[rgb]{0.00,0.50,0.00}{##1}}}
\@namedef{PYG@tok@m}{\def\PYG@tc##1{\textcolor[rgb]{0.40,0.40,0.40}{##1}}}
\@namedef{PYG@tok@gh}{\let\PYG@bf=\textbf\def\PYG@tc##1{\textcolor[rgb]{0.00,0.00,0.50}{##1}}}
\@namedef{PYG@tok@gu}{\let\PYG@bf=\textbf\def\PYG@tc##1{\textcolor[rgb]{0.50,0.00,0.50}{##1}}}
\@namedef{PYG@tok@gd}{\def\PYG@tc##1{\textcolor[rgb]{0.63,0.00,0.00}{##1}}}
\@namedef{PYG@tok@gi}{\def\PYG@tc##1{\textcolor[rgb]{0.00,0.52,0.00}{##1}}}
\@namedef{PYG@tok@gr}{\def\PYG@tc##1{\textcolor[rgb]{0.89,0.00,0.00}{##1}}}
\@namedef{PYG@tok@ge}{\let\PYG@it=\textit}
\@namedef{PYG@tok@gs}{\let\PYG@bf=\textbf}
\@namedef{PYG@tok@ges}{\let\PYG@bf=\textbf\let\PYG@it=\textit}
\@namedef{PYG@tok@gp}{\let\PYG@bf=\textbf\def\PYG@tc##1{\textcolor[rgb]{0.00,0.00,0.50}{##1}}}
\@namedef{PYG@tok@go}{\def\PYG@tc##1{\textcolor[rgb]{0.44,0.44,0.44}{##1}}}
\@namedef{PYG@tok@gt}{\def\PYG@tc##1{\textcolor[rgb]{0.00,0.27,0.87}{##1}}}
\@namedef{PYG@tok@err}{\def\PYG@bc##1{{\setlength{\fboxsep}{\string -\fboxrule}\fcolorbox[rgb]{1.00,0.00,0.00}{1,1,1}{\strut ##1}}}}
\@namedef{PYG@tok@kc}{\let\PYG@bf=\textbf\def\PYG@tc##1{\textcolor[rgb]{0.00,0.50,0.00}{##1}}}
\@namedef{PYG@tok@kd}{\let\PYG@bf=\textbf\def\PYG@tc##1{\textcolor[rgb]{0.00,0.50,0.00}{##1}}}
\@namedef{PYG@tok@kn}{\let\PYG@bf=\textbf\def\PYG@tc##1{\textcolor[rgb]{0.00,0.50,0.00}{##1}}}
\@namedef{PYG@tok@kr}{\let\PYG@bf=\textbf\def\PYG@tc##1{\textcolor[rgb]{0.00,0.50,0.00}{##1}}}
\@namedef{PYG@tok@bp}{\def\PYG@tc##1{\textcolor[rgb]{0.00,0.50,0.00}{##1}}}
\@namedef{PYG@tok@fm}{\def\PYG@tc##1{\textcolor[rgb]{0.00,0.00,1.00}{##1}}}
\@namedef{PYG@tok@vc}{\def\PYG@tc##1{\textcolor[rgb]{0.10,0.09,0.49}{##1}}}
\@namedef{PYG@tok@vg}{\def\PYG@tc##1{\textcolor[rgb]{0.10,0.09,0.49}{##1}}}
\@namedef{PYG@tok@vi}{\def\PYG@tc##1{\textcolor[rgb]{0.10,0.09,0.49}{##1}}}
\@namedef{PYG@tok@vm}{\def\PYG@tc##1{\textcolor[rgb]{0.10,0.09,0.49}{##1}}}
\@namedef{PYG@tok@sa}{\def\PYG@tc##1{\textcolor[rgb]{0.73,0.13,0.13}{##1}}}
\@namedef{PYG@tok@sb}{\def\PYG@tc##1{\textcolor[rgb]{0.73,0.13,0.13}{##1}}}
\@namedef{PYG@tok@sc}{\def\PYG@tc##1{\textcolor[rgb]{0.73,0.13,0.13}{##1}}}
\@namedef{PYG@tok@dl}{\def\PYG@tc##1{\textcolor[rgb]{0.73,0.13,0.13}{##1}}}
\@namedef{PYG@tok@s2}{\def\PYG@tc##1{\textcolor[rgb]{0.73,0.13,0.13}{##1}}}
\@namedef{PYG@tok@sh}{\def\PYG@tc##1{\textcolor[rgb]{0.73,0.13,0.13}{##1}}}
\@namedef{PYG@tok@s1}{\def\PYG@tc##1{\textcolor[rgb]{0.73,0.13,0.13}{##1}}}
\@namedef{PYG@tok@mb}{\def\PYG@tc##1{\textcolor[rgb]{0.40,0.40,0.40}{##1}}}
\@namedef{PYG@tok@mf}{\def\PYG@tc##1{\textcolor[rgb]{0.40,0.40,0.40}{##1}}}
\@namedef{PYG@tok@mh}{\def\PYG@tc##1{\textcolor[rgb]{0.40,0.40,0.40}{##1}}}
\@namedef{PYG@tok@mi}{\def\PYG@tc##1{\textcolor[rgb]{0.40,0.40,0.40}{##1}}}
\@namedef{PYG@tok@il}{\def\PYG@tc##1{\textcolor[rgb]{0.40,0.40,0.40}{##1}}}
\@namedef{PYG@tok@mo}{\def\PYG@tc##1{\textcolor[rgb]{0.40,0.40,0.40}{##1}}}
\@namedef{PYG@tok@ch}{\let\PYG@it=\textit\def\PYG@tc##1{\textcolor[rgb]{0.24,0.48,0.48}{##1}}}
\@namedef{PYG@tok@cm}{\let\PYG@it=\textit\def\PYG@tc##1{\textcolor[rgb]{0.24,0.48,0.48}{##1}}}
\@namedef{PYG@tok@cpf}{\let\PYG@it=\textit\def\PYG@tc##1{\textcolor[rgb]{0.24,0.48,0.48}{##1}}}
\@namedef{PYG@tok@c1}{\let\PYG@it=\textit\def\PYG@tc##1{\textcolor[rgb]{0.24,0.48,0.48}{##1}}}
\@namedef{PYG@tok@cs}{\let\PYG@it=\textit\def\PYG@tc##1{\textcolor[rgb]{0.24,0.48,0.48}{##1}}}

\def\PYGZbs{\char`\\}
\def\PYGZus{\char`\_}
\def\PYGZob{\char`\{}
\def\PYGZcb{\char`\}}
\def\PYGZca{\char`\^}

\def\PYGZlt{\char`\<}
\def\PYGZgt{\char`\>}
\def\PYGZsh{\char`\#}

\def\PYGZdl{\char`\$}
\def\PYGZhy{\char`\-}
\def\PYGZsq{\char`\'}

\makeatother
\fvset{tabcolor=black}
\begingroup
\catcode`\'=\active
\gdef\PYGZsq{'}
\endgroup
\def\PYGZhy{\mbox{-}}
\VerbatimPygments{\PYG}{\PYG}
\newcommand{\fcLeanInline}[2][]{\UseVerb[breaklines,breakbefore={. },breakafter={_},#1]{#2}}

\definecolor{keywordcolor}{rgb}{0.7, 0.1, 0.1}
\definecolor{tacticcolor}{rgb}{0.0, 0.1, 0.6}
\definecolor{commentcolor}{rgb}{0.4, 0.4, 0.4}
\definecolor{symbolcolor}{rgb}{0.0, 0.1, 0.6}
\definecolor{sortcolor}{rgb}{0.1, 0.5, 0.1}
\definecolor{attributecolor}{rgb}{0.7, 0.1, 0.1}

  \definecolor{diffstart}{named}{Grey}
  \definecolor{diffincl}{named}{Green}
  \definecolor{diffrem}{named}{OrangeRed}
  \lstdefinelanguage{diff}{
    basicstyle=\ttfamily\small,
    morecomment=[f][\color{diffstart}]{@@},
    morecomment=[f][\color{diffincl}]{+ },
    morecomment=[f][\color{diffrem}]{- },
  }

\title{Formal Conjectures: An Open and Evolving Benchmark for Verified Discovery in Mathematics}

\author{%
  \hspace*{-\tabcolsep}\textbf{Moritz Firsching}$^1$\thanks{Equal contribution. Correspondence to: \texttt{firsching@google.com}.} \hspace{6pt}%
  \textbf{Paul Lezeau}$^{2, 3}$\footnotemark[1] \thanks{Work partially done at Google DeepMind. \quad $^{\ddagger}$Work done at Google DeepMind.}  \hspace{6pt}%
  \textbf{Salvatore Mercuri}$^{2}$\footnotemark[1] \footnotemark[2] \hspace{6pt}%
  \textbf{Mikl\'{o}s Z.~Horv\'{a}th}$^1$\footnotemark[1]\hspace*{-\tabcolsep} \\
  \textbf{Ya\"{e}l Dillies}$^4$\footnotemark[3] \quad
  \textbf{Calle S\"{o}nne}$^{2, 3}$\footnotemark[3] \quad
  \textbf{Eric Wieser}$^1$ \quad
  \textbf{Fred Zhang}$^1$ \\
  \textbf{Thomas Hubert}$^1$ \quad
  \textbf{Blaise Ag\"{u}era y Arcas}$^5$ \quad
  \textbf{Pushmeet Kohli}$^1$ \\
  \vspace{0.1cm} \\
  $^1$Google DeepMind \quad $^2$Imperial College London \\ $^3$London School of Geometry and Number Theory \\
  $^4$Stockholms universitet \quad $^5$Google
}

\begin{document}
\SaveVerb[commandchars=\\\{\}]{fcLean0}{\PYG{o}{(}\PYG{l+m+mi}{0}\PYG{+w}{ }\PYG{o}{:}\PYG{+w}{ }\PYG{n}{ℕ}\PYG{o}{)}}%
\SaveVerb[commandchars=\\\{\}]{fcLean1}{\PYG{o}{(}\PYG{l+m+mi}{1}\PYG{+w}{ }\PYG{o}{:}\PYG{+w}{ }\PYG{n}{ℚ}\PYG{o}{)}}%
\SaveVerb[commandchars=\\\{\}]{fcLean2}{\PYG{n+nb+bp}{.}\PYG{n}{answer}}%
\SaveVerb[commandchars=\\\{\}]{fcLean3}{\PYG{l+m+mi}{1}\PYG{+w}{ }\PYG{n+nb+bp}{/}\PYG{+w}{ }\PYG{l+m+mi}{2}\PYG{+w}{ }\PYG{n+nb+bp}{*}\PYG{+w}{ }\PYG{l+m+mi}{2}\PYG{+w}{ }\PYG{n+nb+bp}{=}\PYG{+w}{ }\PYG{n}{foo}\PYG{n+nb+bp}{.}\PYG{n}{answer}}%
\SaveVerb[commandchars=\\\{\}]{fcLean4}{\PYG{l+m+mi}{1}\PYG{+w}{ }\PYG{n+nb+bp}{/}\PYG{+w}{ }\PYG{l+m+mi}{2}\PYG{+w}{ }\PYG{n+nb+bp}{*}\PYG{+w}{ }\PYG{l+m+mi}{2}\PYG{+w}{ }\PYG{n+nb+bp}{=}\PYG{+w}{ }\PYG{g+gr}{sorry}}%
\SaveVerb[commandchars=\\\{\}]{fcLean5}{\PYG{l+m+mi}{42}}%
\SaveVerb[commandchars=\\\{\}]{fcLean6}{\PYG{o}{:=}\PYG{+w}{ }\PYG{g+gr}{sorry}}%
\SaveVerb[commandchars=\\\{\}]{fcLean7}{\PYG{n}{False}}%
\SaveVerb[commandchars=\\\{\}]{fcLean8}{\PYG{n}{G}}%
\SaveVerb[commandchars=\\\{\}]{fcLean9}{\PYG{n}{Partition}}%
\SaveVerb[commandchars=\\\{\}]{fcLean10}{\PYG{k+kt}{Prop}}%
\SaveVerb[commandchars=\\\{\}]{fcLean11}{\PYG{n}{True}}%
\SaveVerb[commandchars=\\\{\}]{fcLean12}{\PYG{n}{V}}%
\SaveVerb[commandchars=\\\{\}]{fcLean13}{\PYG{n}{answer}\PYG{o}{(}\PYG{+w}{ }\PYG{o}{)}}%
\SaveVerb[commandchars=\\\{\}]{fcLean14}{\PYG{n}{answer}\PYG{o}{(}\PYG{g+gr}{sorry}\PYG{o}{)}}%
\SaveVerb[commandchars=\\\{\}]{fcLean15}{\PYG{n}{answer}\PYG{o}{(}\PYG{g+gr}{sorry}\PYG{o}{)}\PYG{+w}{ }\PYG{n+nb+bp}{↔}\PYG{+w}{ }\PYG{n}{P}}%
\SaveVerb[commandchars=\\\{\}]{fcLean16}{\PYG{k+kn}{def}\PYG{+w}{ }\PYG{n}{foo}\PYG{n+nb+bp}{.}\PYG{n}{answer}\PYG{+w}{ }\PYG{o}{:}\PYG{+w}{ }\PYG{n}{ℕ}\PYG{+w}{ }\PYG{o}{:=}\PYG{+w}{ }\PYG{g+gr}{sorry}}%
\SaveVerb[commandchars=\\\{\}]{fcLean17}{\PYG{k+kn}{def}\PYG{+w}{ }\PYG{n}{foo}\PYG{n+nb+bp}{.}\PYG{n}{answer}\PYG{+w}{ }\PYG{o}{:}\PYG{+w}{ }\PYG{n}{ℚ}\PYG{+w}{ }\PYG{o}{:=}\PYG{+w}{ }\PYG{g+gr}{sorry}}%
\SaveVerb[commandchars=\\\{\}]{fcLean18}{\PYG{n}{foo}\PYG{n+nb+bp}{.}\PYG{n}{answer}}%
\SaveVerb[commandchars=\\\{\}]{fcLean19}{\PYG{n}{formal\PYGZus{}conjectures}}%
\SaveVerb[commandchars=\\\{\}]{fcLean20}{\PYG{n}{research}\PYG{+w}{ }\PYG{n}{solved}}%
\SaveVerb[commandchars=\\\{\}]{fcLean21}{\PYG{n}{smallest\PYGZus{}n}\PYG{n+nb+bp}{.}\PYG{n}{answer}}%
\SaveVerb[commandchars=\\\{\}]{fcLean22}{\PYG{g+gr}{sorry}}%
\SaveVerb[commandchars=\\\{\}]{fcLean23}{\PYG{n}{sup}}%
\SaveVerb[commandchars=\\\{\}]{fcLean24}{\PYG{k+kn}{theorem}\PYG{+w}{ }\PYG{n}{foo}}%
\SaveVerb[commandchars=\\\{\}]{fcLean25}{\PYG{n+nb+bp}{↔}\PYG{+w}{ }\PYG{n}{answer}\PYG{o}{(}\PYG{g+gr}{sorry}\PYG{o}{)}}%

\maketitle

\setcounter{footnote}{0}

\newcommand{\fcCountTotal}{7559}
\newcommand{\fcCountResearchOpen}{1642}
\newcommand{\fcCountResearchSolved}{2923}
\newcommand{\fcCountTest}{2144}
\newcommand{\fcCountAPI}{608}
\newcommand{\fcCountTextbook}{242}
\newcommand{\fcCountSolvedGroup}{3165}
\newcommand{\fcCountInfraGroup}{2752}
\newcommand{\fcCountAllOther}{2994}
\newcommand{\fcCountInfraPct}{36.4}
\newcommand{\fcSrcErdos}{3424}
\newcommand{\fcSrcWikipedia}{879}
\newcommand{\fcSrcGreens}{284}
\newcommand{\fcSrcPapers}{360}
\newcommand{\fcSrcOQP}{127}
\newcommand{\fcSrcWOWII}{171}
\newcommand{\fcSrcOEIS}{1800}
\newcommand{\fcSrcArxiv}{240}
\newcommand{\fcSrcMO}{68}
\newcommand{\fcSrcOther}{206}
\newcommand{\fcOverviewSourceRows}{%
        Erd\H{o}s Problems       & \textbf{\fcSrcErdos} \\
        OEIS                     & \textbf{\fcSrcOEIS} \\
        Wikipedia                & \textbf{\fcSrcWikipedia} \\
        Papers                   & \textbf{\fcSrcPapers} \\
        Green's Open Problems    & \textbf{\fcSrcGreens} \\
        arXiv                    & \textbf{\fcSrcArxiv} \\
        Written on the Wall II   & \textbf{\fcSrcWOWII} \\
        Open Quantum Problems    & \textbf{\fcSrcOQP} \\
        MathOverflow             & \textbf{\fcSrcMO} \\
        Other sources            & \textbf{\fcSrcOther} \\
}
\newcommand{\fcAMSQuantum}{175}
\newcommand{\fcAMSLinearAlgebra}{190}
\newcommand{\fcAMSAlgGeom}{180}
\newcommand{\fcAMSConvexGeom}{371}
\newcommand{\fcProofSolvedWithProofPct}{42.2}
\newcommand{\fcProofSolvedInRepoPct}{5.1}
\newcommand{\fcProofSolvedLinkedPct}{37.1}
\newcommand{\fcCountForMathlibFiles}{251}

\begin{abstract}
As automated reasoning systems advance rapidly, there is a growing need for research-level formal mathematical problems to accurately evaluate their capabilities.
To address this, we present Formal Conjectures,\footnote{\url{https://github.com/google-deepmind/formal-conjectures}} an evolving benchmark of currently \fcCountTotal{} mathematical problem statements formalized in Lean~4.
Sourced from areas of active mathematical research, the dataset features \fcCountResearchOpen{} open research conjectures without known solutions for mathematical proof discovery, \fcCountResearchSolved{} solved problems for proof auto-formalization, and \fcCountAllOther{} supporting statements.
Notably, the repository provides a structured interface connecting mathematicians who formalize and clarify problems with the AI systems and humans attempting to solve them.
Demonstrating its immediate utility, the benchmark has already been leveraged for new mathematical discoveries, including the resolution of open research conjectures, and for the formal confirmation of major newly announced results.
We describe our approach to ensuring the correctness of these formalizations in a collaborative open-source project where contributions stem from an active community.
In this framework, AI-generated proofs and disproofs serve as a valuable auditing mechanism to iteratively improve the fidelity of the benchmark.
Finally, we provide a standardized evaluation setup and report baseline results on frozen evaluation subsets, demonstrating a climbable signal that measures the current frontier of automated reasoning on research-level mathematics.
\end{abstract}

\section{Introduction}\label{sec:introduction}

As automated reasoning systems, including large language models (LLMs), continue improving their capabilities, the benchmarks used to measure their progress face significant challenges.
In the domain of (formal) mathematics, many existing evaluation frameworks suffer from: \textbf{(i) Data leakage:} As solutions to benchmark problems appear online, it is difficult to distinguish genuine reasoning from memorization. This is a challenge for benchmarks like MiniF2F, PutnamBench, FATE, and FormalMATH \citep{minif2f21, putnambench24, jiang2025fate, yu2025formalmath}; \textbf{(ii) Secretive evaluation:} To combat leakage, some benchmarks keep the evaluation set private, which hinders open and reproducible research \citep{ARC24, frontiermath24, humanitysexam25}; \textbf{(iii) Oversimplified success criteria:} Many benchmarks rely on verifying a simple, machine-checkable final answer (such as a number), which fails to capture the complex, multi-step reasoning inherent in a full mathematical proof \citep{hendrycks2measuring, frontiermath24}; and \textbf{(iv) Saturation:} As models improve, some benchmarks become saturated. For instance, MiniF2F \citep{minif2f21} is now routinely solved at over 99\% accuracy \citep{AlphaProof25, chen2025seed}, with open-weight provers close behind \citep{lin2025goedelproverv2}.

To address these limitations, we introduce \textbf{Formal Conjectures}, an open-source and growing collection of mathematical problems formalized in Lean~4 \citep{moura2021lean} using Mathlib \citep{mathlib2020}.
The benchmark focuses on \emph{open conjectures} as the primary challenge, providing a testbed where complete solutions are not known in the literature while a conjecture remains open. The secondary challenge consists of (informally) solved problems for auto-formalization, where informal proofs are already known.
We formalize all problems in a formal language to allow for objective, rigorous, and automated evaluation: a proposed solution is definitively correct if and only if its corresponding proof is accepted by the kernel without relying on forbidden axioms like \lstinline{sorry}.
Our choice of Lean~4 is pragmatic: Mathlib, the largest formal mathematics library available, is essential for stating a wide variety of advanced conjectures.
While we focus on Lean~4, the core principles of our benchmark are language-agnostic and extendable to other formal systems.

This paper details the construction and methodology of the \textbf{Formal Conjectures} benchmark.
The benchmark is publicly available under the Apache~2.0 and CC-BY~4.0 licenses and has already been utilized for real-world mathematical discovery \citep{alexeev25, DMProverAgent26}. In summary, our main contributions are:
\begin{itemize}
    \item \textbf{The Formal Conjectures Benchmark:} A large-scale, evolving dataset of \fcCountTotal{} mathematical statements in Lean~4, comprising \fcCountResearchOpen{} open conjectures without known solutions for proof discovery, \fcCountResearchSolved{} solved problems for auto-formalization, and \fcCountAllOther{} supporting textbook, API, and test statements.
    \item \textbf{A Unified API for Mathematical Reasoning:} A repository interface connecting mathematicians with an ecosystem of automated solvers to facilitate simultaneous statement auditing and rapid dissemination of problems to AI systems.
    \item \textbf{Methodology for Statement Formalization:} We introduce a collaborative pipeline and a three-level taxonomy of misformalizations (Translation, Underspecified, and Source) to ensure formalization fidelity. This includes the \lstinline{@[question]} mechanism, which decouples mathematical value discovery from proof verification. We provide insights from 291 fixed cases of research-level formalization.
    \item \textbf{A Standardized Evaluation Framework:} We provide a rigorous evaluation setup using the Lean~4 kernel to verify multi-step reasoning without relying on oversimplified numerical answers. We establish frozen subsets, \texttt{FC100SolvedSet1} and \texttt{FC100OpenSet1}, to enable stable, reproducible model comparisons.
\end{itemize}

\section{Related Work}\label{sec:related_work}

Our work intersects mathematical reasoning benchmarks and formal theorem proving.

\paragraph{General mathematical reasoning benchmarks.}
Several benchmarks evaluate AI systems on advanced problem-solving without formal proofs.
FrontierMath \citep{frontiermath24} and Humanity's Last Exam \citep{humanitysexam25} rely on expert-crafted questions with machine-checkable final answers that do not capture full multi-step proofs, while MathArena \citep{matharena25} evaluates LLMs on fresh competition problems to reduce contamination risk and assesses informal proof-writing.
Following our initial May 2025 release, both FrontierMath and MathArena added research-level counterparts: Epoch AI launched \emph{FrontierMath: Open Problems} (a January 2026 website-only pilot offering evaluation as a service), and MathArena \citep{dekoninck2026beyond} expanded to research problems, including a Lean track.
Further research-level benchmarks followed: HorizonMath \citep{wang2026measuring} automatically verifies candidate answers to 113 predominantly unsolved problems in computational and applied mathematics, while IMProofBench \citep{schmitt2025improofbench} and First Proof \citep{abouzaid2026first} (some of whose problems are formalized in our dataset) evaluate research-level informal proof generation.
The ARC Prize \citep{ARC24, ARC25} targets abstract reasoning, sharing our focus on measuring genuine reasoning over pattern matching.
Formal benchmarks, by contrast, allow automatic verification of full reasoning or proof.

\paragraph{Formal mathematics benchmarks and automated theorem proving.}
Progress in AI reasoning and automated theorem proving has been mutually reinforcing, leading to the rapid saturation of existing formal benchmarks.
MiniF2F \citep{minif2f21} pioneered high-school-level evaluation but became increasingly saturated as models improved.
This shift was marked by AlphaProof \citep{AlphaProof25} and AlphaGeometry 2 \citep{chervonyi2025gold} reaching silver-medal performance at the 2024 IMO, where AlphaProof solved Lean problems and introduced the formal-imo \citep{AlphaProof25} dataset.
Systems like Aristotle \citep{achim2025aristotle} and Seed-Prover \citep{chen2025seed} achieved gold-medal equivalents at the 2025 IMO.
This capability leap cascaded to undergraduate math, where benchmarks like PutnamBench \citep{putnambench24} and Putnam 2025 are being saturated by systems such as Seed-Prover 1.5 \citep{chen2025seed15}, Numina \citep{liu2026numina}, and others.
While SorryDB \citep{sorrydb26} evaluates AI on practical theorem completion and FATE \citep{jiang2025fate} targets varying difficulties of frontier algebra, the saturation of earlier benchmarks highlights an acute need for research-level evaluation: benchmarking reasoning on unsolved research-level mathematics.

\begin{figure}[t!]
    \centering
    \resizebox{\textwidth}{!}{
        \begin{tikzpicture}[font=\sffamily]
  \definecolor{fcSrcDark}{RGB}{9, 115, 86}
  \definecolor{fcSrcBorder}{RGB}{106, 194, 166}
  \definecolor{fcSrcBg}{RGB}{241, 250, 246}

  \definecolor{fcBenchDark}{RGB}{24, 78, 158}
  \definecolor{fcBenchBorder}{RGB}{122, 166, 224}
  \definecolor{fcBenchBg}{RGB}{241, 246, 253}

  \definecolor{fcOpenAccent}{RGB}{6, 140, 96}
  \definecolor{fcSolvedAccent}{RGB}{28, 96, 194}
  \definecolor{fcSuppAccent}{RGB}{82, 78, 166}

  \definecolor{fcPipeAccent}{RGB}{198, 78, 18}
  \definecolor{fcTextMain}{RGB}{22, 30, 42}
  \definecolor{fcTextSub}{RGB}{78, 90, 106}
  \definecolor{fcCardRule}{RGB}{214, 223, 234}

  \tikzset{
    boxsrc/.style={
      fill=fcSrcBg,
      draw=fcSrcBorder,
      line width=0.95pt,
      rounded corners=5.5pt
    },
    boxbench/.style={
      fill=fcBenchBg,
      draw=fcBenchBorder,
      line width=0.95pt,
      rounded corners=5.5pt
    },
    srcitem/.style={
      rectangle,
      draw=fcSrcBorder!55!fcCardRule,
      fill=white,
      line width=0.55pt,
      rounded corners=4pt,
      text width=4.48cm,
      inner xsep=6.5pt,
      inner ysep=3.2pt,
      align=left,
      font=\sffamily\fontsize{6.3}{7.05}\selectfont,
      text=fcTextMain
    },
    repoitem/.style={
      rectangle,
      draw=#1!38!fcCardRule,
      fill=white,
      line width=0.55pt,
      rounded corners=4pt,
      text width=5.64cm,
      inner xsep=8.0pt,
      inner ysep=3.1pt,
      align=left,
      font=\sffamily\fontsize{6.35}{7.2}\selectfont,
      text=fcTextMain,
      path picture={
        \fill[#1] (path picture bounding box.south west)
          rectangle ([xshift=2.9pt]path picture bounding box.north west);
      }
    },
    pipearrow/.style={
      line width=1.85pt,
      color=fcPipeAccent,
      -{Stealth[length=6.6pt, width=5.6pt]}
    }
  }

  \node (repo_open) [repoitem=fcOpenAccent] at (7.50, 0) {%
    {\fontsize{7.9}{8.75}\selectfont\bfseries Research Open\strut}%
    \hfill
    {\fontsize{8.1}{8.75}\selectfont\bfseries\color{fcOpenAccent!90!black}\fcCountResearchOpen\strut}\\[0.5pt]
    {\fontsize{6.6}{7.4}\selectfont\color{fcTextMain!92}$\hookrightarrow$ Primary Goal: \textbf{New Proof Discovery}\strut}\\[0.3pt]
    {\fontsize{6.05}{6.85}\selectfont\color{fcTextSub}Unsolved conjectures with no known solutions.\strut}%
  };

  \node (repo_solved) [repoitem=fcSolvedAccent, below=0.095cm of repo_open] {%
    {\fontsize{7.9}{8.75}\selectfont\bfseries Research Solved\strut}%
    \hfill
    {\fontsize{8.1}{8.75}\selectfont\bfseries\color{fcSolvedAccent!90!black}\fcCountResearchSolved\strut}\\[0.5pt]
    {\fontsize{6.6}{7.4}\selectfont\color{fcTextMain!92}$\hookrightarrow$ Secondary Goal: \textbf{Proof Auto-formalization}\strut}\\[0.3pt]
    {\fontsize{6.05}{6.85}\selectfont\color{fcTextSub}Established theorems with known informal proofs.\strut}%
  };

  \node (repo_api) [repoitem=fcSuppAccent, below=0.095cm of repo_solved] {%
    {\fontsize{7.9}{8.75}\selectfont\bfseries Supporting Statements\strut}%
    \hfill
    {\fontsize{8.1}{8.75}\selectfont\bfseries\color{fcSuppAccent!92!black}\fcCountAllOther\strut}\\[0.6pt]
    {\fontsize{6.35}{7.15}\selectfont
      {\color{fcSuppAccent}$\bullet$}~\textbf{Test:}~\fcCountTest\hfill
      {\color{fcSuppAccent}$\bullet$}~\textbf{API:}~\fcCountAPI\hfill
      {\color{fcSuppAccent}$\bullet$}~\textbf{Textbook:}~\fcCountTextbook\strut%
    }\\[0.35pt]
    {\fontsize{6.6}{7.4}\selectfont\color{fcTextMain!92}$\hookrightarrow$ Purpose: \textbf{Definition Testing \& Reusable API}\strut}%
  };

  \node (src_all) [
    srcitem,
    left=2.18cm of repo_open.north west,
    anchor=north east
  ] {%
    {\fontsize{6.1}{6.85}\selectfont\bfseries\color{fcSrcDark!92!black}Source Collection\hfill \# Statements\strut}\\[1.1pt]
    {\color{fcCardRule}\hrule height 0.5pt}\vspace{1.5pt}%
    {\fontsize{6.35}{7.15}\selectfont\renewcommand{\arraystretch}{1.055}%
    \begin{tabular*}{\linewidth}{@{\extracolsep{\fill}}l@{\hspace{4pt}}r@{}}
      \fcOverviewSourceRows
    \end{tabular*}}\\[1.5pt]
    {\color{fcCardRule}\hrule height 0.45pt}\vspace{1.4pt}%
    {\fontsize{5.55}{6.35}\selectfont\color{fcTextSub}Archives \& problem lists (incl.\ variants \& tests).\strut\par}%
  };

  \node (source_title) [
    font=\sffamily\fontsize{8.15}{9.2}\selectfont\bfseries,
    text=fcSrcDark,
    above=0.11cm of src_all.north
  ] {Informal Sources\strut};

  \node (repo_title) [
    font=\sffamily\fontsize{8.15}{9.2}\selectfont\bfseries,
    text=fcBenchDark,
    above=0.11cm of repo_open.north
  ] {Formal Conjectures Benchmark\strut};

  \begin{scope}[on background layer]
    \node (bg_inputs) [
      boxsrc,
      fit=(source_title) (src_all) (src_all.west |- repo_api.south),
      inner sep=4.0pt
    ] {};
    \node (bg_repo) [
      boxbench,
      fit=(repo_title) (repo_open) (repo_api) (repo_api.east |- src_all.south),
      inner sep=4.0pt
    ] {};
  \end{scope}

  \path (bg_inputs.east) -- (bg_repo.west) coordinate[midway] (mid_gap);

  \draw [pipearrow]
    ([xshift=5.2pt]bg_inputs.east |- mid_gap) --
    ([xshift=-5.2pt]bg_repo.west |- mid_gap);

  \node [
    font=\sffamily\fontsize{7.2}{8.2}\selectfont\bfseries,
    text=fcPipeAccent,
    align=center,
    above=3.0pt of mid_gap
  ] {Formalization\\Pipeline};

  \node [
    font=\sffamily\fontsize{6.0}{7.0}\selectfont\bfseries,
    text=fcTextSub!92!black,
    align=center,
    below=3.2pt of mid_gap
  ] {Lean~4, Mathlib\\\& Metadata};

\end{tikzpicture}%
    }
    \vspace{-10pt}
    \caption{Left: informal source distribution. Right: formal category distribution, including open (\fcCountResearchOpen) and solved (\fcCountResearchSolved) problems.}
    \label{fig:benchmark_pipeline}
\end{figure}

\paragraph{Adoption of Formal Conjectures.}\label{sec:impact}
Since its open-source release in May 2025, the repository has been adopted by both mathematicians and tool builders.
Boris Alexeev used a repository formalization to prove Erd\H{o}s Problem 124 \citep{erdos124} with Aristotle \citep{achim2025aristotle}; although the statement initially included an unintended extra hypothesis, Aristotle's proof succeeded without relying on it, and subsequent works have resolved further repository problems with AI-assisted proofs \citep{alexeev25, guo2026greatest}, the latter settling an OEIS conjecture of Stephan that is part of \texttt{FC100OpenSet1}.
Crucially, because problem statements are formalized and vetted in advance, the repository has also served as a trusted reference target to confirm major newly claimed results: following informal announcements of a counterexample to the Jacobian conjecture in dimension $n \ge 3$ \citep{TaoJacobian26} and of a complex structure on $S^6$ resolving the Hopf problem \citep{AlpogeS6_26}, formal proofs against the pre-existing statements in Formal Conjectures confirmed both results (the Jacobian counterexample directly in the repository and the Hopf problem in \citet{AlexeevHopf26}).
Similarly, OpenAI announced a proof of finite-time blowup for the forced 3D Navier--Stokes equations, establishing two of the four alternatives (C and D) in the official statement of the Clay Millennium Prize Problem \citep{fefferman2006existence}; the accompanying Lean~4 formalization is our repository's formal statement, adapted only in imports, attributes, namespace, and notation \citep{OpenAINavierStokes26}.
In addition, AlphaProof Nexus \citep{DMProverAgent26} conducted a systematic evaluation of the full repository, autonomously resolving several open problems, including nine Erd\H{o}s problems.
To facilitate standardized, reproducible evaluation, this paper introduces named, versioned snapshots (Section~\ref{sec:experimental_setup}), moving beyond individual tests to systematic benchmarking across diverse methods.

\section{Formal Conjectures}\label{sec:formal_conjectures}

\subsection{Problem Selection and Composition}

The benchmark's problems are sourced from diverse mathematical literature for broad coverage.
Key sources include the collection of problems posed by Paul Erd\H{o}s, as cataloged on \emph{erdosproblems.com} \citep{erdosproblems}; the On-Line Encyclopedia of Integer Sequences (OEIS); well-known Wikipedia conjectures; conjectures from recent publications in academic journals and on arXiv; Ben Green's list of open problems; Written on the Wall~II; the IQOQI Vienna list of open quantum problems \citep{oqp}; the Kourovka Notebook of unsolved problems in group theory \citep{kourovka}; and questions asked on MathOverflow.
The number of problems per informal source is in Figure \ref{fig:benchmark_pipeline}.
Furthermore, the repository has grown steadily since its initial open-source release, as visualized in Figure \ref{fig:repo_growth}.

\input{sections/assets/growth.tex}

The collection is divided into two primary categories: \textbf{Unsolved Conjectures}, open research problems where a formal or informal proof would represent a new mathematical discovery; and \textbf{Solved Problems}, established theorems with known \emph{informal} proofs that benchmark an AI's ability to formalize existing arguments, though only a fraction currently have a \emph{formal} proof.
We include both categories not only because they independently serve as benchmarks for different tasks (proof generation and auto-formalization), but also because they are often intertwined.
When formalizing an open conjecture, it is natural to also state and prove simpler, solved variations.
For instance, a general conjecture about the existence of Hadamard matrices of size $4k$ for all valid $k \in \mathbb{N}$ can be presented alongside the solved cases (e.g., for all $k \le 166$) and the first open case ($k = 167$).

Other sources of \emph{solved} problems are collections of conjectures that include both open and solved questions. In those cases, we also formalize the statements of the solved cases. They are often related in subject to the open conjectures, and hence the formalization of their statements and the auto-formalization of their proofs support the open conjectures.

Alongside research-level solved problems, the repository also includes simpler supporting statements (\lstinline[language={}]{test}, \lstinline[language={}]{API}, and \lstinline[language={}]{textbook}), ranging from trivial sanity checks to graduate textbook results, which help validate the correctness of the underlying formalizations.
Including such simpler problems helps detect incorrect formalizations of open problems: a disproof of an easy special case or known variant might indicate that some of the definitions used have been stated incorrectly.
See Section~\ref{sec:misform} and Appendix~\ref{sec:app_misformalization_examples} for examples of how those initial misformalizations can lead to improved formal and informal problem statements.
We strive to always state the simplest open variant of a conjecture, as well as the conjecture in greater generality.

\paragraph{The \lstinline{@[category]} attribute.}
To help navigate this collection, each statement is labeled with one of the following category tags: \lstinline[language={}]{research open} (an open research-level problem), \lstinline[language={}]{research solved} (a solved research-level problem), \lstinline[language={}]{textbook} (a textbook-level math problem ranging from high school to graduate level), \lstinline[language={}]{API} (statements developing foundational theory for a new definition intended for general reuse), and \lstinline[language={}]{test} (a statement serving as a sanity check).
The distribution of all categories is in Figure~\ref{fig:benchmark_pipeline} (note that the project website hides \lstinline[language={}]{test} and \lstinline[language={}]{API} statements by default, whereas the total of \fcCountTotal{} reported here counts all tagged statements).

\paragraph{The \lstinline{@[AMS]} attribute.}
Statements are also labeled with subject tags following the AMS Mathematics Subject Classification \citep{ams2020}, written as \lstinline{AMS} followed by the subject codes (e.g., \lstinline{AMS 14} in Figure~\ref{fig:example-statement}).
The distribution by source collection and AMS subject classification is shown in Tables~\ref{tab:problems-by-collection} and~\ref{tab:ams-breakdown} (Appendix~\ref{sec:app_source_ams_tables}).
While number theory and combinatorics account for over half of all statements, other areas have a significant presence across over 10 AMS subjects (e.g., \fcAMSConvexGeom{} statements in convex and discrete geometry, \fcAMSLinearAlgebra{} in linear and multilinear algebra, \fcAMSAlgGeom{} in algebraic geometry, and \fcAMSQuantum{} in quantum theory), which we aim to broaden as the repository grows.

\subsection{Formalization Methodology}\label{sec:formalization-methodology}

\begin{figure}[t!]
    \centering
\begin{Verbatim}[commandchars=\\\{\}]
\PYG{l+s+sd}{/\PYGZhy{}\PYGZhy{}}
\PYG{l+s+sd}{There are no indecomposable vector bundles of rank 2 on \PYGZdl{}\PYGZbs{}mathbb\PYGZob{}P\PYGZcb{}\PYGZca{}n\PYGZdl{}}
\PYG{l+s+sd}{for \PYGZdl{}n \PYGZbs{}ge 7\PYGZdl{}.}
\PYG{l+s+sd}{\PYGZhy{}/}
\PYG{k+kd}{@[}\PYG{n}{category}\PYG{+w}{ }\PYG{n}{research}\PYG{+w}{ }\PYG{n}{open}\PYG{o}{,}\PYG{+w}{ }\PYG{n}{AMS}\PYG{+w}{ }\PYG{l+m+mi}{14}\PYG{k+kd}{]}
\PYG{k+kn}{theorem}\PYG{+w}{ }\PYG{n}{hartshorne\PYGZus{}conjecture}\PYG{+w}{ }\PYG{o}{(}\PYG{n}{n}\PYG{+w}{ }\PYG{o}{:}\PYG{+w}{ }\PYG{n}{ℕ}\PYG{o}{)}\PYG{+w}{ }\PYG{o}{(}\PYG{n}{hn}\PYG{+w}{ }\PYG{o}{:}\PYG{+w}{ }\PYG{l+m+mi}{7}\PYG{+w}{ }\PYG{n+nb+bp}{≤}\PYG{+w}{ }\PYG{n}{n}\PYG{o}{)}
\PYG{+w}{    }\PYG{o}{(}\PYG{n}{F}\PYG{+w}{ }\PYG{o}{:}\PYG{+w}{ }\PYG{n}{VectorBundles}\PYG{+w}{ }\PYG{o}{(}\PYG{n}{ProjectiveSpace}\PYG{+w}{ }\PYG{o}{(}\PYG{n}{Fin}\PYG{+w}{ }\PYG{o}{(}\PYG{n}{n}\PYG{+w}{ }\PYG{n+nb+bp}{+}\PYG{+w}{ }\PYG{l+m+mi}{1}\PYG{o}{))}\PYG{+w}{ }\PYG{o}{(}\PYG{n}{Spec}\PYG{+w}{ }\PYG{o}{(}\PYG{n+nb+bp}{.}\PYG{n}{of}\PYG{+w}{ }\PYG{n}{ℂ}\PYG{o}{))))}
\PYG{+w}{    }\PYG{o}{(}\PYG{n}{hF}\PYG{+w}{ }\PYG{o}{:}\PYG{+w}{ }\PYG{n}{F}\PYG{n+nb+bp}{.}\PYG{n}{rank}\PYG{+w}{ }\PYG{n+nb+bp}{=}\PYG{+w}{ }\PYG{l+m+mi}{2}\PYG{o}{)}\PYG{+w}{ }\PYG{o}{:}\PYG{+w}{ }\PYG{n}{Nonempty}\PYG{+w}{ }\PYG{o}{(}\PYG{n}{F}\PYG{n+nb+bp}{.}\PYG{n}{Splitting}\PYG{+w}{ }\PYG{o}{(}\PYG{n}{Fin}\PYG{+w}{ }\PYG{l+m+mi}{2}\PYG{o}{))}\PYG{+w}{ }\PYG{o}{:=}\PYG{+w}{ }\PYG{k}{by}
\PYG{+w}{  }\PYG{g+gr}{sorry}
\end{Verbatim}
    \caption{A statement in Formal Conjectures, an open conjecture from \citet{hartshorne1974varieties}. It includes an informal statement as a comment, tags, and a Lean statement.}
    \label{fig:example-statement}
\end{figure}

All problems are presented as Lean theorems with informal statements (see Figure~\ref{fig:example-statement}). While our core goal is providing open statements with \lstinline{sorry} placeholders for AI solvers to replace, we also include solved variants of conjectures and items from problem lists. For open problems, no proofs are provided. For solved items, we appreciate very short proofs or counterexamples (under 25--50 lines) that test the definitions. Longer proofs are excluded to keep the repository lightweight; instead, contributors are encouraged to host proofs in their own repositories and link to them using the \lstinline{@[formal_proof]} mechanism described below.

\paragraph{The \texorpdfstring{\lstinline{@[question]}}{@[question]} attribute.}
Many mathematical problems ask not just for a proof but for a specific \emph{answer}: a number, a function, or a set.
To handle this, we introduce a \lstinline{@[question value]} attribute paired with a companion answer definition, separating the task of discovering an answer from proving it correct.
For example, \enquote{What is the smallest $n$ such that $P(n)$?} is formalized with a companion placeholder (e.g., \fcLeanInline{fcLean16}) referenced as \fcLeanInline{fcLean18} in \fcLeanInline{fcLean24}, which a solver must replace with a concrete value, reducing the problem to a verifiable proof obligation.
Similarly, yes-or-no (\fcLeanInline{fcLean10}-valued) questions whose truth value is unknown state $P$ directly (or $\neg P$ when resolved negatively) under the \lstinline{@[question]} attribute.
This attribute-based design replaced an earlier \fcLeanInline{fcLean14} elaborator (used in \texttt{bench-v1}) to better support external verifiers (Appendix~\ref{sec:app_answer_sorry}).

\paragraph{The \enquote{for Mathlib} pattern.}
The project maintains a \texttt{FormalConjecturesForMathlib} directory containing \fcCountForMathlibFiles{} files of auxiliary definitions and lemmas not yet in Mathlib but needed for conjectures.
These results are asynchronously contributed upstream to Mathlib.
See Appendix~\ref{sec:app_for_mathlib} for details on the rationale and contents of this directory.

\paragraph{The \lstinline{@[formal_proof]} attribute.}
To keep the repository lightweight, the \lstinline{@[formal_proof]} attribute decouples problem statements from their solutions.
It supports three modes: (1)~\fcLeanInline{fcLean19} for proofs solving exactly the type of the statement given here; it should point to a commit on top of a commit in the \texttt{main} branch of our repo, filling in the \fcLeanInline{fcLean22};
(2)~\lstinline{lean4} for equivalent problems solved in Lean~4 elsewhere (e.g., Mathlib or external repos); and
(3)~\lstinline{other_system} for problems solved in other systems like Lean~3, Isabelle, or Rocq.
This links to external verification sources while keeping the repository maintainable.
The attribute applies to any solved category, though usage on a \lstinline{research open} problem triggers a linter warning.
Table~\ref{tab:formal_proof_baseline} summarizes proof coverage recorded via this attribute.

\subsection{Misformalizations}
\label{sec:misform}

It is notoriously difficult to formalize a mathematical problem \emph{correctly} without providing a formal proof.
A \emph{misformalization} is a formal statement that does not faithfully capture the intended mathematical statement.
To provide a rigorous foundation for benchmark quality, we introduce a taxonomy of misformalizations.
It includes the following levels, where in all cases the formal statement is unfaithful.
\begin{enumerate}
    \item \textbf{Translation}: the informal statement is accurate and explicitly phrased.
    \item \textbf{Underspecified}: the informal statement is accurate but lacks detail.
    \item \textbf{Source}: the informal statement is not as intended.
\end{enumerate}

These levels are ordered with respect to the degree of contribution from the formalizer.
Level 1 misformalizations arise from the formalization only, while level 3 misformalizations arise from the informal text only.
Moreover, for a Lean expert who does not necessarily have domain expertise in the informal statements, levels are ordered with respect to difficulty to fix.
Note that level 2 and 3 misformalizations have been useful in clarifying informal statements in the literature.
An illustrative example is Erd\H{o}s Problem~978: multiple attempts to formalize the statement led to a refinement of the original informal problem.\footnote{\url{https://www.erdosproblems.com/forum/thread/978}}
The original statement,
\enquote{Are there infinitely many $n$ for which $f(n)$ is $(k-2)$-power-free?},
was found to require multiple additional hypotheses and became:
\enquote{If $k>3$, $k$ is not a power of two, and for all primes $p$ there exists $n$ such that $p^{k-2}\nmid f(n)$, then are there infinitely many $n$ for which $f(n)$ is $(k-2)$-power-free?}
More broadly, informal texts may exclude trivial cases without explicitly describing them, while formalization requires these to be explicit.

Misformalizations can further be categorized into six types across the three levels: \emph{syntactic}, \emph{semantic}, and \emph{misrepresentation} errors at the translation level; \emph{implicit conventions} at the underspecified level; and \emph{reporting} and \emph{mathematical} errors at the source level.
The full taxonomy with definitions and representative pull requests is given in Appendix~\ref{sec:app_misform_taxonomy}.
Across the repository a total of $291$ misformalizations have been fixed, with \emph{misrepresentation} (48\%) and \emph{semantic} (35\%) errors being the most common (Table~\ref{tab:misform_table}).
Detailed code diffs illustrating these misformalization examples can be found in Appendix~\ref{sec:app_misformalization_examples}.

\subsection{Avoiding Misformalizations}
\label{sec:avoiding_misform}

\begin{figure}[t!]
    \centering
    \resizebox{\textwidth}{!}{%
        \begin{tikzpicture}[
  node distance=1.02cm and 1.30cm,
  font=\sffamily
]
  \definecolor{fcSrcDark}{RGB}{9, 115, 86}
  \definecolor{fcSrcBorder}{RGB}{106, 194, 166}
  \definecolor{fcSrcBg}{RGB}{241, 250, 246}

  \definecolor{fcFormDark}{RGB}{198, 78, 18}
  \definecolor{fcFormBorder}{RGB}{234, 156, 104}
  \definecolor{fcFormBg}{RGB}{255, 247, 238}

  \definecolor{fcRepoDark}{RGB}{24, 78, 158}
  \definecolor{fcRepoBorder}{RGB}{122, 166, 224}
  \definecolor{fcRepoBg}{RGB}{241, 246, 253}

  \definecolor{fcRunDark}{RGB}{126, 34, 148}
  \definecolor{fcRunBorder}{RGB}{190, 136, 208}
  \definecolor{fcRunBg}{RGB}{250, 243, 253}

  \definecolor{fcAuditDark}{RGB}{82, 78, 166}
  \definecolor{fcAuditBorder}{RGB}{148, 144, 214}
  \definecolor{fcAuditBg}{RGB}{244, 243, 253}

  \definecolor{fcOutDark}{RGB}{11, 116, 126}
  \definecolor{fcOutBorder}{RGB}{102, 184, 192}
  \definecolor{fcOutBg}{RGB}{239, 250, 251}

  \definecolor{fcFixRed}{RGB}{190, 42, 38}
  \definecolor{fcSrcSlate}{RGB}{92, 72, 108}
  \definecolor{fcValidGreen}{RGB}{6, 134, 92}
  \definecolor{fcTextSub}{RGB}{76, 88, 104}
  \definecolor{fcArrowSlate}{RGB}{88, 100, 118}

  \tikzset{
    basebox/.style={
      rectangle,
      line width=1.0pt,
      rounded corners=5pt,
      text width=3.35cm,
      minimum height=0.98cm,
      inner xsep=5.5pt,
      inner ysep=4.2pt,
      align=center,
      font=\sffamily\fontsize{6.8}{7.9}\selectfont
    },
    flowsrc/.style={basebox, fill=fcSrcBg, draw=fcSrcBorder},
    flowform/.style={basebox, fill=fcFormBg, draw=fcFormBorder},
    flowrepo/.style={basebox, fill=fcRepoBg, draw=fcRepoBorder},
    flowrun/.style={basebox, fill=fcRunBg, draw=fcRunBorder},
    flowaudit/.style={basebox, fill=fcAuditBg, draw=fcAuditBorder},
    flowout/.style={basebox, fill=fcOutBg, draw=fcOutBorder},
    fwdarrow/.style={
      line width=1.3pt,
      color=fcArrowSlate,
      -{Stealth[length=6.0pt, width=5.0pt]},
      shorten >=1.5pt,
      shorten <=1.5pt
    },
    runarrow/.style={
      line width=1.3pt,
      color=fcRunDark!85!fcArrowSlate,
      -{Stealth[length=6.0pt, width=5.0pt]},
      shorten >=1.5pt,
      shorten <=1.5pt
    },
    fixarrow/.style={
      line width=1.25pt,
      color=fcFixRed,
      densely dashed,
      -{Stealth[length=5.8pt, width=4.8pt]},
      shorten >=1.5pt,
      shorten <=1.5pt
    },
    srcarrow/.style={
      line width=1.25pt,
      color=fcSrcSlate,
      densely dashed,
      -{Stealth[length=5.8pt, width=4.8pt]},
      shorten >=1.5pt,
      shorten <=1.5pt
    },
    validarrow/.style={
      line width=1.35pt,
      color=fcValidGreen,
      -{Stealth[length=6.0pt, width=5.0pt]},
      shorten >=1.5pt,
      shorten <=1.5pt
    },
    pathlabel/.style={
      font=\sffamily\fontsize{6.7}{7.6}\selectfont\bfseries,
      align=center,
      inner sep=1.2pt
    }
  }

  \node (source) [flowsrc] {%
    {\fontsize{8.3}{9.6}\selectfont\bfseries\color{fcSrcDark}Informal Source\strut}\\[1.4pt]
    {\fontsize{6.7}{7.8}\selectfont\color{fcTextSub}Erd\H{o}s, OEIS, arXiv, papers\strut}%
  };

  \node (formalize) [flowform, right=of source] {%
    {\fontsize{8.3}{9.6}\selectfont\bfseries\color{fcFormDark}Formalization\strut}\\[1.4pt]
    {\fontsize{6.7}{7.8}\selectfont\color{fcTextSub}Human/AI Lean~4 \& tests\strut}%
  };

  \node (repo) [flowrepo, right=of formalize] {%
    {\fontsize{8.3}{9.6}\selectfont\bfseries\color{fcRepoDark}Merge to Repository\strut}\\[1.4pt]
    {\fontsize{6.7}{7.8}\selectfont\color{fcTextSub}PR review \& Lean linters\strut}%
  };

  \node (eval) [flowrun, below=of repo] {%
    {\fontsize{8.3}{9.6}\selectfont\bfseries\color{fcRunDark}Automated Runs\strut}\\[1.4pt]
    {\fontsize{6.7}{7.8}\selectfont\color{fcTextSub}AlphaProof \& AI provers\strut}%
  };

  \node (inspect) [flowaudit, below=of formalize] {%
    {\fontsize{8.3}{9.6}\selectfont\bfseries\color{fcAuditDark}Verification \& Triage\strut}\\[1.4pt]
    {\fontsize{6.7}{7.8}\selectfont\color{fcTextSub}Kernel \& fidelity audit\strut}%
  };

  \node (status) [flowout, below=of source] {%
    {\fontsize{8.3}{9.6}\selectfont\bfseries\color{fcOutDark}Status Updated\strut}\\[1.4pt]
    {\fontsize{6.7}{7.8}\selectfont\color{fcTextSub}In repo \& informal source\strut}%
  };

  \draw [fwdarrow] (source.east) -- (formalize.west);
  \draw [fwdarrow] (formalize.east) -- (repo.west);
  \draw [fwdarrow] (repo.south) -- (eval.north);

  \draw [runarrow] (eval.west) -- (inspect.east)
    node[midway, above=1.8pt, pathlabel, text=fcRunDark] {Proofs \&\\[-0.5pt]Disproofs};

  \draw [fixarrow] (inspect.north) -- (formalize.south)
    node[midway, right=3.2pt, pathlabel, align=left, text=fcFixRed] {%
      Misformalization\\[-0.5pt]%
      {\sffamily\mdseries\fontsize{5.85}{6.8}\selectfont\color{fcFixRed!88!black}(Translation / Underspec.)}%
    };

  \draw [srcarrow] (inspect.north west) -- (source.south east)
    node[pos=0.52, above=1.2pt, sloped, font=\sffamily\fontsize{6.45}{7.3}\selectfont\bfseries, inner sep=1pt, text=fcSrcSlate!95!black] {Source Error};

  \draw [validarrow] (inspect.west) -- (status.east)
    node[midway, above=1.8pt, pathlabel, text=fcValidGreen!95!black] {Valid\\[-0.5pt]Result};

\end{tikzpicture}%
%
    }%
    \caption{Iterative pipeline for Formal Conjectures. Formalized statements are tested by periodic AI runs, solving problems or triggering a revision loop.}
    \label{fig:formalization_process}
\end{figure}

In Formal Conjectures, every contribution undergoes mandatory code review by humans with both Lean expertise and relevant mathematical domain knowledge.
Prior to its initial May 2025 release, the general setup and initial problem set also underwent an independent audit by core Mathlib maintainers and external formalization experts.
While early contributions to the repository were written entirely by hand, the authoring process has evolved: most recent submissions use AI tooling, including agentic coding assistants and auto-formalization systems.
To facilitate high-quality AI-assisted contributions, the repository provides an \texttt{AGENTS.md} file with structured guidance for these tools.
Similarly, the review process increasingly leverages AI: reviewers use language models to cross-check formalizations against informal sources, and AlphaProof runs on submissions to catch potential misformalizations before merging.

Beyond this review process, we employ a number of additional strategies to mitigate the risk of misformalizations across the dataset.
Firstly, we employ and encourage a test-based design for definitions.
Any new definitions introduced should be accompanied by a suite of proven \texttt{test} lemmas and \texttt{API} statements that are designed to establish
expected behavior of the definition and mitigate the risk of edge-case errors.
Lean and \lstinline{Mathlib}'s tooling (for example \lstinline{decide}) can be employed to prove many of these \texttt{test} statements.
Currently, the repository contains \fcCountTest{} \texttt{test} statements and \fcCountAPI{} \texttt{API} statements, amounting to \fcCountInfraPct\% of the repository statements.

Secondly, automated theorem provers like AlphaProof regularly attempt proofs and disproofs across the repository (Figure \ref{fig:formalization_process}); manually inspecting these results primarily reveals misformalizations.
In contrast to test-based design, which proactively reduces the risk of edge-case errors, this retroactive approach can unearth misformalizations across all categories as described in Section~\ref{sec:misform}.
In addition, we use Gemini, AlphaProof, and other tools to automatically cross-check formalizations against their informal source texts, flagging potential discrepancies for human review.
Future work involves further automating these processes in GitHub CI to reduce the overhead of manual post-hoc fixes by maintainers.

Thirdly, custom Lean linters built on Mathlib's framework enforce metadata and documentation standards across contributions.
These cover AMS classification tags, \lstinline{@[question]} usage, problem category annotations, and module docstrings.

Finally, as Formal Conjectures has grown in stature, ongoing community audits by domain mathematicians, including experienced contributors who now review new submissions under our review guidelines, have been increasingly valuable alongside proof and disproof attempts from external automated theorem provers.
Furthermore, insights and corrections discovered through this process are actively upstreamed to the original sources, e.g., prompting the maintainer of the Erdős problems website to regularly clarify and adjust the informal problem statements.

\section{Benchmark Evaluation}\label{sec:experimental_evaluation}

\subsection{Evaluation Framework}\label{sec:experimental_setup}

\paragraph{Evaluation Paradigms.}
Formal Conjectures provides a dynamic, rigorous benchmark for advanced automated reasoning premised on novel mathematical insight as the ultimate intellectual hurdle for complex problems. We define two distinct setups:
\begin{enumerate}
    \item \textbf{Primary Goal: New Formal Proof Discovery.}
    Measuring AI's ability to discover formal proofs for open problems, this track targets unsolved conjectures with no known formal or informal solutions. Prior to a conjecture being solved, no solutions exist in any training corpus, so the first proof provides a strong signal of genuine mathematical discovery and reduces the reliance on secretive evaluation sets common in other benchmarks.
    \item \textbf{Secondary Goal: Proof Auto-formalization.}
    This track provides a climbable benchmark treating the non-trivial translation of established mathematics into formal code as a distinct, rigorous challenge. It measures a model's ability to work with Lean~4 and Mathlib to formalize known arguments (informal proofs exist in the literature but are not necessarily provided to the solver in our baselines).
\end{enumerate}

\paragraph{Evaluating Open Problems.}
Evaluating open problems is clear-cut: because no formal or informal solutions are known while a conjecture remains open, the first kernel-accepted proof provides an uncontaminated success signal, after which the problem is reclassified as \texttt{research solved}.
However, since a proof may settle a misformalized statement instead of the intended one, any result for an open conjecture triggers manual inspection for fidelity.

\paragraph{Fixed Benchmark Subsets.}
To enable stable model comparisons, we provide two frozen subsets of 100 problems each:
\texttt{FC100OpenSet1} (100 \texttt{research open} statements) and \texttt{FC100SolvedSet1} (100 statements sampled from all non-open categories: 50 \texttt{research solved}, 34 \texttt{test}, 9 \texttt{API}, 7 \texttt{textbook}).
These subsets are defined in the repository files \texttt{FC100OpenSet1.lean} and \texttt{FC100SolvedSet1.lean}, which import exactly the corresponding theorem statements.
Because these files are compiled as part of the repository across all supported Lean versions, the problem sets remain fixed and comparable even as the repository evolves.
Further details on the construction of these subsets—as well as newly released, curated second sets (\texttt{FC100OpenSet2} and \texttt{FC100SolvedSet2}) selected for mathematical impact, formalization confidence, and the absence of existing formal proofs—are in Appendix~\ref{sec:app_bench_lists}.

\paragraph{Correctness.}
Following established formal math benchmark methodologies \citep{minif2f21, putnambench24}, our evaluation leverages the Lean~4 kernel for an objective, rigorous, and automatable success criterion.
A proposed solution is a Lean proof term replacing the \fcLeanInline{fcLean22} placeholder (together with a value for the companion \fcLeanInline{fcLean2} definition, or \fcLeanInline{fcLean14} in \texttt{bench-v1}, where applicable).
Solutions are correct if and only if accepted by the kernel using at most the standard Lean~4 axioms (\texttt{propext}, \texttt{Classical.choice}, and \texttt{Quot.sound}), excluding forbidden axioms such as \lstinline{sorry} (\texttt{sorryAx}) and \texttt{Lean.ofReduceBool} (\texttt{native\_decide}).
This binary criterion eliminates human grading ambiguity and provides an indisputable ground truth.
While theoretically susceptible to foundational bugs in the Lean kernel or unintended axioms (Appendix~\ref{sec:illustrative_evaluation}), in practice, kernel-level verification provides the highest mathematical rigor.
To guard against models exploiting environment modifications (such as redefining local instances) to satisfy the kernel without proving the intended statement, we combine kernel verification with standalone tools such as \texttt{SafeVerify} \citep{safeverify25} and \texttt{Comparator} \citep{comparator25}, alongside manual inspection of discovered solutions, allowing verified proof terms to be shared publicly for transparent, reproducible research.

\paragraph{Versioning and Reproducibility.}\label{sec:versioning}
Evaluating on a living repository is challenging: dependency updates can alter statement semantics, and new Mathlib theorems may simplify problems.
To enable reproducible comparisons, we tag frozen benchmark snapshots using a two-part naming scheme: \texttt{bench-v\textit{N}-lean4.\textit{X}.\textit{Y}}.
The \emph{bench version} \texttt{v\textit{N}} identifies a fixed problem set; the \emph{Lean version} suffix pins the Mathlib tag (and hence Lean toolchain) against which statements compile and are evaluated.
Releases are issued every few months; when Lean versions bump, companion tags allow evaluating the same problem set against intermediate toolchains, e.g., \texttt{bench-v1-lean4.29.0}.
To preserve baselines, snapshots are immutable; misformalization fixes are never patched retroactively, only into \texttt{bench-v(N+1)}.

\paragraph{A Living Benchmark and Unified API.}
Outside the frozen subsets, the repository operates as a dynamic benchmark that exhibits useful evaluation dynamics over time.
As new conjectures are added, solved, or corrected by the community, it naturally self-adjusts its difficulty, expands, and improves.
The hardest problems remain open, pushing the frontier of automated reasoning.
Crucially, the repository acts as a unified API: researchers can submit formalized conjectures to a single centralized hub for simultaneous exposure to all evaluating AI systems.
This obviates the need to manually engage with individual provers, ensuring broad and concurrent attempts at mathematical discovery.

\begin{table}[t!]
  \caption{Proof coverage across non-open statement categories. \emph{Proved in repo}: the statement has a \lstinline{sorry}-free proof in the repository; \emph{Linked proof}: a proof is recorded via the \lstinline{@[formal\_proof]} attribute but is not in the repository itself; \emph{With proof}: either of the above.}
  \label{tab:formal_proof_baseline}
  \centering
  \begin{tabular}{lrrrr}
    \toprule
    Category & Total & \multicolumn{3}{c}{Proof Status} \\
    \cmidrule(lr){3-5}
    & & Proved in repo & Linked proof & With proof \\
    \midrule
    \texttt{research solved} & 2923 & 150 (5.1\%) & 1084 (37.1\%) & 1234 (42.2\%) \\
    \texttt{textbook} & 242 & 132 (54.5\%) & 26 (10.7\%) & 158 (65.3\%) \\
    \texttt{test} & 2144 & 2103 (98.1\%) & 6 (0.3\%) & 2109 (98.4\%) \\
    \texttt{API} & 608 & 596 (98.0\%) & 2 (0.3\%) & 598 (98.4\%) \\
    \midrule
    \textbf{Total} & 5917 & 2981 (50.4\%) & 1118 (18.9\%) & 4099 (69.3\%) \\
    \bottomrule
  \end{tabular}
\end{table}

\paragraph{Open Problems Baseline.}
For the primary proof-discovery task on open problems, the baseline for the first tagged release \href{https://github.com/google-deepmind/formal-conjectures/releases/tag/bench-v1-lean4.27.0}{\texttt{bench-v1-lean4.27.0}} is $0\%$ for all systems: by definition, no formal proof exists for any \texttt{research open} statement at the time of tagging (problems with known solutions are reclassified as \texttt{research solved}).
Over time, these open statements may receive formal or informal proofs; indeed, since our initial release, several open conjectures across the repository have already been resolved by automated theorem-proving systems and community efforts (see Section~\ref{sec:impact}).
Notably, since tagging, 17 of the 100 \texttt{FC100OpenSet1} statements have been reclassified as solved (13 with a formal proof, in the repository or linked externally; 1 as a consequence of a newly resolved related conjecture; and 3 after discovering they had already been settled in the literature).
Proofs that expose a misformalization (Section~\ref{sec:misform}) rather than settling the intended problem trigger revisions incorporated into the next version.
In subsequent releases, interim-solved problems are reclassified to \texttt{research solved}, with formal solutions receiving the \lstinline{@[formal_proof]} attribute.
Across the full repository, the updated \texttt{research open} pool (and newly curated subsets such as \texttt{FC100OpenSet2}) combines remaining unsolved problems with new additions, starting again at a $0\%$ baseline, whereas fixed subsets such as \texttt{FC100OpenSet1} retain all 100 original members across versions for longitudinal tracking (Appendix~\ref{sec:app_bench_lists}).

\paragraph{Proof Auto-formalization Baseline.}
For auto-formalization, the baseline is system-agnostic: we track all formal proofs contributed to the repository, regardless of origin.
These are tracked using the \lstinline{@[formal_proof]} attribute (Section~\ref{sec:formalization-methodology}), which records the proof system and links to the source.
Table~\ref{tab:formal_proof_baseline} summarizes the current state.
Notably, \fcProofSolvedWithProofPct\% of \texttt{research solved} statements have a known formal proof (\fcProofSolvedInRepoPct\% internal, \fcProofSolvedLinkedPct\% linked externally).

\subsection{Illustrative Evaluation}\label{sec:experimental_results}

\paragraph{Frozen Evaluation Subsets.}
We provide illustrative evaluations, with setup, prompting, and cost details in Appendix~\ref{sec:illustrative_evaluation}.
To ensure stable comparisons despite repository fluidity, we provide two frozen 100-problem subsets from \texttt{bench-v1-lean4.27.0}: \texttt{FC100SolvedSet1} (proof auto-formalization) and \texttt{FC100OpenSet1} (proof discovery; $0\%$ baseline).
Our initial modest-budget runs on \texttt{FC100OpenSet1} (Gemini~3.1~Pro, AlphaProof, and AlphaProof Nexus) solved $0\%$, highlighting its difficulty as a frontier benchmark for mathematical discovery, though larger-scale agentic runs and community efforts have since resolved several open conjectures across the repository and \texttt{FC100OpenSet1} (see Section~\ref{sec:impact} and above).
We encourage the community to report results on both frozen subsets and the evolving main repository.

\begin{wraptable}{r}{0.45\textwidth}
  \vspace{-15pt}
  \caption{Results on the frozen \texttt{FC100SolvedSet1} (mean $\pm$ SD over 10 runs where applicable).}
  \vspace{5pt}
  \label{tab:results_bench_table}
  \centering
  \small
  \begin{tabular}{lr@{ }l}
    \toprule
    Method & \multicolumn{2}{c}{Proved (\%)} \\
    \midrule
    Gemini 3.1 Pro (0-shot) & $22.2$ & $\pm\, 1.2$ \\
    GPT 5.6 Sol (0-shot) & $28.6$ & $\pm\, 1.1$ \\
    GPT 5.6 Terra (0-shot) & $25.8$ & $\pm\, 1.9$ \\
    GPT 5.6 Luna (0-shot) & $26.5$ & $\pm\, 1.0$ \\
    GPT 5.6 Luna (tool use) & $62.0$ & \\
    Claude Sonnet 5 (0-shot) & $27.7$ & $\pm\, 1.3$ \\
    Claude Sonnet 5 (tool use) & $59.0$ & \\
    Claude Opus 5 (0-shot) & $38.6$ & $\pm\, 2.1$ \\
    Claude Opus 5 (tool use) & $68.0$ & \\
    Leanstral 1.5 (tool use) & $38.0$ & \\
    AlphaProof (1k sims) & $43.1$ & $\pm\, 1.2$ \\
    AlphaProof (16k sims) & $48.7$ & $\pm\, 0.9$ \\
    AlphaProof Nexus (dev) & $66.0$ & \\
    \bottomrule
  \end{tabular}
  \vspace{-10pt}
\end{wraptable}

\paragraph{Baseline Results on \texttt{FC100SolvedSet1}.}
Results on \texttt{FC100SolvedSet1} (Table~\ref{tab:results_bench_table}) demonstrate a consistent, climbable signal across model capabilities, tool use, and search compute, with low variance across multi-run evaluations ($\text{SD}$ of $0.9\%$--$2.1\%$; multi-turn tool-use and agentic systems are evaluated in a single run).
We evaluate four distinct paradigms: (1)~single-turn 0-shot LLMs; (2)~multi-turn LLMs with Lean tools; (3)~a slightly improved version of the standalone AlphaProof reinforcement-learning tree-search system used in the 2024 IMO \citep{AlphaProof25}; and (4)~a development version of AlphaProof Nexus \citep{DMProverAgent26}, an agentic framework running Gemini~3.1~Pro in a multi-turn loop with Lean compiler feedback, evolutionary sketch selection, and AlphaProof subgoal calls.
In 0-shot, mean solve rates demonstrate scaling with base model capability, spanning $22.2\%$ for Gemini~3.1~Pro, $25.8\%$--$28.6\%$ across GPT~5.6 (Terra, Luna, Sol) and Claude Sonnet~5 ($27.7\%$), and $38.6\%$ for Claude Opus~5.
Enabling Lean tool use yields substantial gains over 0-shot: GPT~5.6~Luna jumps from $26.5\%$ to $62.0\%$, and Claude Sonnet~5 and Opus~5 reach $59.0\%$ and $68.0\%$, while open-weight Leanstral~1.5 \citep{Leanstral26} (6B active parameters) reaches $38.0\%$.
Standalone AlphaProof (3B parameters; tree-search inference only on v6e TPUs, without test-time RL) demonstrates inference-time search scaling: averaged across ten independent single-attempt runs, solve rates rise from $43.1\%$ at 1,000 simulations to $48.7\%$ at 16,000 simulations ($0\%$ disproofs across all runs).
Moreover, AlphaProof Nexus (dev) reaches $66.0\%$, outperforming both its Gemini~3.1~Pro backbone ($22.2\%$) and standalone AlphaProof ($48.7\%$) by combining agentic sketching with specialist subgoal search.
Finally, since systems prove the formal Lean statements as written, misformalizations affect fidelity to the informal source rather than formal validity.
A post-release audit of all 100 statements in \texttt{FC100SolvedSet1@bench-v1-lean4.27.0} (combining repository fixes with manual and LLM-assisted checks) identified 14 unfaithful statements ($14\%$ lower bound): 7 involve minor semantic shifts that remain valid problems of at least equal difficulty, 4 are unprovable as stated (capping the solve rate at $96\%$), and only 3 weaken the intended claim and risk false positives (at most 2 via trivial shortcuts), confirming minimal impact on Table~\ref{tab:results_bench_table}.

\section{Discussion}\label{sec:discussion}

\paragraph{Broader Benefits.}
Beyond its primary role, Formal Conjectures provides \textbf{precise mathematics} by enforcing a level of formal precision that clarifies the exact meaning of statements.
This creates a library of conjectures referenced unambiguously via unique file names and commit hashes, ensuring fixed and verifiable mathematical intent, unlike standard citations where generality or precision is often left to interpretation.
Furthermore, the benchmark serves as a compass for \textbf{Mathlib development}, as formalizing advanced conjectures reveals gaps in the existing library.
We address these via \texttt{FormalConjecturesForMathlib}, a directory of results continuously upstreamed to Mathlib. By identifying hard-to-state concepts, the repository both benchmarks reasoning and expands formalized mathematics.

\paragraph{Limitations.}
Problem selection faces a \textbf{scope bias} constrained by Mathlib's current coverage, which we mitigate somewhat via our \texttt{ForMathlib} directory; this limitation will diminish as Mathlib expands.
There is also a \textbf{selection bias} toward famous conjectures and Erd\H{o}s problems, concentrating on number theory and combinatorics; we plan to mitigate this by sourcing diverse arXiv preprints in areas like computer science and physics.
Focusing on open problems introduces a \textbf{misformalization risk} from subtle statement or source errors.
Although mitigated by the audits in Section~\ref{sec:avoiding_misform}, these risks remain until a full proof is provided, and fixes require version updates; however, detecting flawed or ill-posed statements is itself a valuable reasoning signal, and every caught error reduces the residual error rate in later releases.
The benchmark also faces \textbf{contamination challenges}: for solved problems, models may retrieve known informal arguments or pre-existing in-repository formal proofs of supporting items in \texttt{FC100SolvedSet1} (even when masked at evaluation time, a channel eliminated in \texttt{FC100SolvedSet2} by selecting only unproven \texttt{research solved} statements) rather than reason from scratch; for open problems, although no complete solutions are known, models may have encountered informal discussions or partial results (mirroring how human mathematicians work), and newly discovered proofs will eventually enter training data.
Our living repository tracks these developments and reclassifies problems accordingly.
A related risk is \textbf{status mislabeling}: a statement tagged \texttt{research open} may already have been settled in the literature without the formalizers being aware; these are reclassified once discovered.
Finally, \textbf{verification and reproducibility} face foundational and practical constraints: verification relies on the Lean~4 kernel's integrity (which, despite theoretical exploitability of foundational bugs or unintended axioms, remains the highest standard for rigor), while reproducing frontier baselines can be limited by proprietary systems and substantial inference compute costs (e.g., dedicated TPU resources or up to \$100 per problem for multi-turn agents).
To support accessible and stable comparisons as the repository evolves, we provide frozen, versioned 100-problem subsets alongside public model baselines.

\paragraph{Future Work.}
As an active open-source project, we are committed to maintaining and extending Formal Conjectures with new problems, improved problems, and platform features; this ongoing commitment is reflected in the repository's steady growth since its initial release (see Figure~\ref{fig:repo_growth}).
See Appendices~\ref{sec:app_bench_lists} and~\ref{sec:extended_discussion} for details on planned extensions, additional frozen evaluation subsets, metadata improvements, and community features.

\section{Conclusion}\label{sec:conclusion}
We present \textbf{Formal Conjectures}, an evolving collection of mathematical problem statements formalized in Lean~4, including open conjectures.
Sourced from a diverse range of mathematical literature and areas, the benchmark provides broad coverage as an evaluation framework for AI systems' capabilities in formal math.
By providing a testbed of open research-level conjectures without known solutions alongside a substantial set for proof auto-formalization, we establish a rigorous environment for evaluating AI for formal mathematical research.
Crucially, to ensure stable and reproducible model comparisons as the repository evolves, we complement this dynamic dataset with versioned, frozen evaluation subsets.
We hope the work can drive the expansion of formalized math: both through its living repository and the continuous upstreaming of new definitions to Mathlib, as well as serve as a high-fidelity benchmark that bridges the gap between human mathematical expertise and automated formal reasoning.
Ultimately, we believe this resource will not only measure current progress but also help accelerate the frontier of formal mathematical discovery.

\begin{ack}
We are grateful to Bhavik Mehta, whose early conceptual vision for the scope and formalization of several research problems was instrumental to this project.
We also thank liudongdong, Daniel Veljanoski, Daniel Chin, Will Blair, Kenta Kitamura, AItoBit, Akhil Mathew, Jack McCarthy, Jean-Guillaume Durand, Yongxi Lin, Felix Pernegger, Seewoo Lee, Codruț Grosu, and Bo Cowgill for their many contributions to the formal-conjectures repository.
More generally, we thank all contributors who have contributed to the repository since it was released as open source, among them
Abel Do\~{n}ate,
Aditya Ramabadran,
Alex Korbonits,
Amogh Parab,
Anirudh Rao,
Anthony Wang,
Arav Bhattacharyya,
Archibald Browne,
Ayush Debnath,
Bolton Bailey,
Chakshu Gupta,
Chris Maki,
Cong Lu,
Daniel Liao,
Derek Steinkamp,
Emmanuel GTZ Rico,
Francesco Nishanil Chotuck,
Franz Huschenbeth,
Hansle Cho,
Himanshu Gupta,
James Jordan,
Jofre Costa,
Joshua Steier,
Junseok Lee,
Junyan Xu,
Kenneth A. Mendoza,
Leandre Jack,
Linmiao Xu,
Madhu Shree Aravindan,
Mario Krenn,
Martin Bruse,
Michael Rothgang,
Mirek Ol\v{s}\'{a}k,
Pratik Jayarao,
Ralf Stephan,
Reklle,
Reza Ramji,
Sunsu Jeong,
The bbchallenge Collaboration (bbchallenge.org),
Victor Li,
Walid Elkersh,
Will Cook,
Wojciech Nawrocki,
Yan Yablonovskiy,
Yoh Tanimoto,
Zeyu Zheng,
and Zhen Ning David Liu.
A full list of contributors is available at \url{https://github.com/google-deepmind/formal-conjectures/graphs/contributors}.
Additionally, we thank Swarat Chaudhuri, George Tsoukalas, Anton Kovsharov, Henryk Michalewski, Edward Lockhart, and Goran Žužić for helpful discussions and support running experiments.
We are also grateful to Johan Commelin, Kevin Buzzard, Oliver Nash, Bhavik Mehta, Floris van Doorn, and Jason Rute for auditing the general setup and initial problem set prior to the public release.
Finally, we thank the anonymous NeurIPS reviewers and Area Chairs for their constructive feedback and suggestions, which significantly improved this paper.
\ 

Paul Lezeau was funded by a London School of Geometry and Number Theory Imperial College London/King’s College London/University College London PhD studentship, which is supported by the Engineering and Physical Sciences Research Council [EP/S021590/1].

Salvatore Mercuri received support through Schmidt Sciences, LLC.

The authors declare no competing financial interests outside their listed affiliations and Google DeepMind internships.
\end{ack}

\bibliographystyle{plainnat}
\bibliography{all}

\appendix

\newpage

\section*{Supplementary Material}

Here, we include various supplementary details.

\section{Formal Conjectures Details}\label{sec:app_formal_conjectures}

\subsection{Source and AMS Classification Tables}\label{sec:app_source_ams_tables}

Note that a single source problem typically gives rise to multiple formal statements due to variants (e.g., special cases, generalizations, solved sub-problems, and test lemmas). We explicitly allow and encourage the inclusion of multiple alternative formalizations for a single informal claim to capture different mathematical perspectives or resolutions of ambiguity; thus, the counts in this table reflect formal statements, not distinct source problems. Moreover, whereas the project website hides \texttt{test} and \texttt{API} lemmas by default, the \fcCountTotal{} total in Table~\ref{tab:problems-by-collection} counts all tagged statements across all five categories.

\begin{table}[h]
    \caption{Number of formal statements by source collection, split by category (\emph{Supporting} aggregates \texttt{test}, \texttt{API}, and \texttt{textbook}). A single source problem may yield multiple statements (variants, special cases, tests).}
    \label{tab:problems-by-collection}
    \centering
    \begin{tabular}{lrrrr}
    \toprule
    \textbf{Source Collection} & \textbf{Total} & \textbf{Research Open} & \textbf{Research Solved} & \textbf{Supporting} \\
    \midrule
    Erd\H{o}s Problems & 3424 & 827 & 1943 & 654 \\
    OEIS & 1800 & 190 & 155 & 1455 \\
    Wikipedia & 879 & 286 & 277 & 316 \\
    Papers & 360 & 97 & 138 & 125 \\
    Green's Open Problems & 284 & 98 & 146 & 40 \\
    arXiv & 240 & 36 & 86 & 118 \\
    Written on the Wall II & 171 & 9 & 54 & 108 \\
    Open Quantum Problems & 127 & 28 & 22 & 77 \\
    MathOverflow & 68 & 9 & 17 & 42 \\
    Other sources & 206 & 62 & 85 & 59 \\
    \midrule
    \textbf{Total} & \textbf{7559} & \textbf{1642} & \textbf{2923} & \textbf{2994} \\
    \bottomrule
    \end{tabular}
\end{table}

Table~\ref{tab:ams-breakdown} shows the distribution of statements across AMS subject classifications.
Other notable areas beyond number theory and combinatorics include convex and discrete geometry, linear algebra, algebraic geometry, and quantum theory.

\begin{table}[h]
    \caption{Top 10 AMS subject classifications by number of statements in Formal Conjectures (statements may carry multiple AMS subject codes, so counts are non-exclusive).}
    \label{tab:ams-breakdown}
    \centering
    \begin{tabular}{lr}
    \toprule
    \textbf{AMS Subject Classification} & \textbf{Statements} \\
    \midrule
    Number theory & 4689 \\
    Combinatorics & 2372 \\
    Convex and discrete geometry & 371 \\
    Linear and multilinear algebra; matrix theory & 190 \\
    Algebraic geometry & 180 \\
    Quantum theory & 175 \\
    Geometry & 141 \\
    Information and communication, circuits & 122 \\
    Mathematical logic and foundations & 120 \\
    Field theory and polynomials & 108 \\
    \bottomrule
    \end{tabular}
\end{table}

\subsection{The \texorpdfstring{\lstinline{@[question]}}{@[question]} Mechanism in Detail}\label{sec:app_answer_sorry}

Problems requiring a solution are formalized using the \lstinline{@[question value]} attribute paired with a companion \fcLeanInline{fcLean2} definition.
Concretely, a problem like \enquote{What is the smallest $n$ such that $P(n)$?} is formalized as:
\par\noindent\begin{minipage}{\linewidth}
\begin{Verbatim}[commandchars=\\\{\}]
\PYG{k+kn}{def}\PYG{+w}{ }\PYG{n}{smallest\PYGZus{}n}\PYG{n+nb+bp}{.}\PYG{n}{answer}\PYG{+w}{ }\PYG{o}{:}\PYG{+w}{ }\PYG{n}{ℕ}\PYG{+w}{ }\PYG{o}{:=}\PYG{+w}{ }\PYG{g+gr}{sorry}
\PYG{k+kd}{@[}\PYG{n}{question}\PYG{+w}{ }\PYG{n}{value}\PYG{k+kd}{]}
\PYG{k+kn}{theorem}\PYG{+w}{ }\PYG{n}{smallest\PYGZus{}n}\PYG{+w}{ }\PYG{o}{:}\PYG{+w}{ }\PYG{n}{IsLeast}\PYG{+w}{ }\PYG{o}{\PYGZob{}}\PYG{n}{n}\PYG{+w}{ }\PYG{o}{:}\PYG{+w}{ }\PYG{n}{ℕ}\PYG{+w}{ }\PYG{n+nb+bp}{|}\PYG{+w}{ }\PYG{n}{P}\PYG{+w}{ }\PYG{n}{n}\PYG{o}{\PYGZcb{}}\PYG{+w}{ }\PYG{n}{smallest\PYGZus{}n}\PYG{n+nb+bp}{.}\PYG{n}{answer}\PYG{+w}{ }\PYG{o}{:=}\PYG{+w}{ }\PYG{k}{by}
\PYG{+w}{  }\PYG{g+gr}{sorry}
\end{Verbatim}
\end{minipage}\par\vspace{\smallskipamount}
As usual, \fcLeanInline{fcLean22} is a placeholder which must be filled in order to claim that a problem is solved, but now there are two of them.
If the solution is known, then \fcLeanInline{fcLean22} in \fcLeanInline[breaklines=false]{fcLean21} is replaced with the value (e.g., \fcLeanInline{fcLean5}).
This mechanism allows our statements to include computational conjectures where discovering the answer is the primary challenge and verification is routine.

Crucially, while providing a proof simply requires finding any way to replace \fcLeanInline{fcLean6} with a valid term, providing an answer is a fundamentally different task: it requires evaluating mathematical meaning to determine the correct value.
Lean's type checker can verify that a proposed answer leads to a valid proof, but it cannot judge whether the answer itself is mathematically meaningful---that remains a job for human mathematicians or AI systems with genuine mathematical reasoning capabilities.

There are a number of advantages to using a companion \fcLeanInline{fcLean2} definition over directly writing a solution or a bare \fcLeanInline{fcLean22} placeholder:
\begin{itemize}
\item It makes clear which part of the formalization corresponds to the solution; this allows the solution to be masked from \fcLeanInline{fcLean20} statements in order to evaluate agents by resetting the \fcLeanInline{fcLean2} definition to \fcLeanInline{fcLean22}, without modifying the theorem statement.
\item
    It provides some protection against Lean's flexible elaboration; a problem stated as \fcLeanInline{fcLean4} could be solved by replacing the \fcLeanInline{fcLean22} with either \fcLeanInline[breaklines=false]{fcLean0} or \fcLeanInline[breaklines=false]{fcLean1},
    as by default, Lean only infers the type of an equality after analyzing both sides.
    This is clearly undesirable; we do not want the meaning of our questions to change depending on the answer provided!
    Declaring \fcLeanInline{fcLean17} and stating the problem instead as \fcLeanInline{fcLean3} protects against this, as the explicit type annotation fixes the type of the equality before seeing the solution.
\item It forces the statement to be written in such a way that does not pre-assume what the solution is.
\end{itemize}

\subsubsection{Propositional Solutions}

A particularly common use case arises when the truth value of a proposition $P$ is itself unknown, i.e., when the problem can be phrased as a question: \enquote{Is it true that\dots?}.
In this case, the statement carries the \lstinline{@[question]} attribute and states $P$ directly:
\begin{Verbatim}[commandchars=\\\{\}]
\PYG{k+kd}{@[}\PYG{n}{question}\PYG{k+kd}{]}
\PYG{k+kn}{theorem}\PYG{+w}{ }\PYG{n}{is\PYGZus{}P\PYGZus{}true}\PYG{+w}{ }\PYG{o}{:}\PYG{+w}{ }\PYG{n}{P}\PYG{+w}{ }\PYG{o}{:=}\PYG{+w}{ }\PYG{k}{by}
\PYG{+w}{  }\PYG{g+gr}{sorry}
\end{Verbatim}
This pattern is widely used throughout the repository for open problems where even the expected truth value is uncertain.
For solver systems that are capable of filling proofs only, stating $P$ directly without a wrapper permits these systems to decide the proposition by attempting to prove either the statement $P$ or its negation $\neg P$ (negating the repository statement to $\neg P$ under \lstinline{@[question]} if resolved negatively).

\subsubsection{Earlier \texorpdfstring{\fcLeanInline{fcLean14}}{answer(sorry)} Representation in \texttt{bench-v1}}

While the \texttt{bench-v1} snapshots and earlier pull requests (such as the historical diffs in Appendix~\ref{sec:app_misformalization_examples}) used a custom Lean term elaborator \fcLeanInline{fcLean13}---writing \fcLeanInline{fcLean14} inline for values and \fcLeanInline{fcLean15} (replaced with \fcLeanInline{fcLean11} or \fcLeanInline{fcLean7}) for yes-or-no questions---we subsequently migrated the main repository to the equivalent declaration-and-attribute representation above to simplify external verification tools such as \texttt{Comparator} and avoid quantifier-scope pitfalls around \fcLeanInline{fcLean25}.\footnote{\url{https://github.com/google-deepmind/formal-conjectures/issues/6690}}

\subsection{The \enquote{for Mathlib} Pattern}\label{sec:app_for_mathlib}

It is typical when undertaking formalization projects to discover missing (and often trivial) results that seem suitable for Mathlib.
Contributing these results to Mathlib is of course the natural resolution, but this has a number of downsides:
\begin{itemize}
\item  New conjectures that depend on these results find themselves at the end of a long dependency chain; the missing results must undergo Mathlib review, the merged change must land in a monthly Mathlib release, and the Formal Conjectures repository must be upgraded to that release.
\item Some results may be too specialized for Mathlib, and a Mathlib contribution would require generalizing these immediately.
\item When the results entail a whole new mathematical development, it may take time for the Mathlib community to find the right abstractions and evolve said development into the right shape. Bypassing the dependency chain mentioned above allows much of this evolution to happen more rapidly outside of Mathlib.
\end{itemize}

The Lean community has widely adopted the pattern of creating a \texttt{ForMathlib} directory as a staging ground for such results, the oldest example of which being \href{https://github.com/leanprover-community/lean-liquid/tree/master/src/for\_mathlib}{the \texttt{for\_mathlib} directory} in the \enquote{liquid tensor experiment}.\footnote{\url{https://github.com/leanprover-community/lean-liquid}}
The formal-conjectures project's \texttt{FormalConjecturesForMathlib} directory follows this pattern, and contains \fcCountForMathlibFiles{} files with results such as natural and logarithmic density of sets of natural numbers (with proofs of basic properties like monotonicity and the density of even numbers), arithmetic progressions and AP-free sets, hypergraph Ramsey numbers, Turing machines and busy beaver halting numbers, perfect powers with a decidability instance, VC dimension in abelian groups, Latin squares and transversals, and a library of graph invariants including the Wiener and Szeged indices, the Lov\'{a}sz theta function, and the Havel--Hakimi residue.
Over time, these results are asynchronously contributed upstream to Mathlib.

\subsection{Misformalization Taxonomy}\label{sec:app_misform_taxonomy}

Misformalizations can be categorized roughly as follows, with the level shown in parentheses.\footnote{Pull request (PR) numbers refer to the repository at \url{https://github.com/google-deepmind/formal-conjectures}; Erd\H{o}s Problem discussions are at \url{https://www.erdosproblems.com}.}
\begin{itemize}
    \item \textbf{Syntactic (1)}: where the formalizer misses a subtlety in the way Lean parses a piece of syntax. For example, this could be missing brackets changing the meaning of an expression in unexpected ways; see PR~\#1338.
    \item \textbf{Semantic (1)}: where the formalizer misses a subtlety in the way that Lean represents types or operators. For example, the natural numbers in Lean contain $0$ and have truncated subtraction; see PRs~\#349, \#1259, \#1260, \#1262.
    \item \textbf{Misrepresentation (1)}: where the formalizer formally misrepresents aspects of the informal statement during translation. For example, the use of incorrect quantifiers, the use of incorrect logical connectives, or the failure to include hypotheses which are present in the source text; see PRs~\#4607, \#1156.
    \item \textbf{Implicit conventions (2)}: where the source text assumes domain expertise and does not explicitly include certain hypotheses or conventions; see PRs~\#3528, \#1151.
    \item \textbf{Reporting (3)}: where the statement changes in the literature through repetition, usually as the result of typos; see Erd\H{o}s Problem~918 discussion.
    \item \textbf{Mathematical (3)}: where the source text is clearly not as intended or contains genuine mathematical errors; see Erd\H{o}s Problem~728 discussion.
\end{itemize}

\begin{table}[H]
    \centering
    \caption{Number and proportion of misformalizations across categories.}
    \label{tab:misform_table}
    \begin{tabular}{llc}
        \textbf{Level} & \textbf{Category} & \textbf{Number of misformalizations} \\
        \toprule
        \multirow{3}{*}{Translation} & Syntactic & 9 (3.09\%)\\
        \cmidrule{2-3}
        & Semantic & 103 (35.40\%)\\
        \cmidrule{2-3}
        & Misrepresentation & 140 (48.11\%) \\
        \midrule
        Underspecified & Implicit conventions & 23 (7.90\%) \\
        \midrule
        \multirow{2}{*}{Source} & Reporting & 15 (5.15\%)\\
        \cmidrule{2-3}
        & Mathematical & 1 (0.34\%) \\
        \bottomrule
    \end{tabular}
\end{table}

\subsection{Misformalization Examples}\label{sec:app_misformalization_examples}

This section contains the misformalization examples mentioned in Section~\ref{sec:misform}. 
Each diff contrasts a misformalized statement with its corrected version. 
The diffs are edited for presentation (using the earlier \fcLeanInline{fcLean14} syntax) and need not match the associated pull requests exactly; some combine several pull requests or internal changes.

\subsubsection{Syntactic Errors}

An example of a syntactic error involving the correction of misplaced parentheses is shown in Figure \ref{fig:misform_syntactic_erdos1054}.

\begin{figure}[h]
    \centering
\begin{Verbatim}[commandchars=\\\{\}]
\PYG{+w}{ } /\PYGZhy{}\PYGZhy{} Let \PYGZdl{}f(n)\PYGZdl{} be the minimal integer \PYGZdl{}m\PYGZdl{} such that \PYGZdl{}n\PYGZdl{} is the sum of the
\PYG{+w}{ } \PYGZdl{}k\PYGZdl{} smallest divisors of \PYGZdl{}m\PYGZdl{} for some \PYGZdl{}k\PYGZbs{}geq 1\PYGZdl{}. Is it true that
\PYG{+w}{ } \PYGZdl{}\PYGZbs{}limsup f(n)/n=\PYGZbs{}infty\PYGZdl{}? \PYGZhy{}/
\PYG{+w}{ } @[category research open, AMS 11]
\PYG{+w}{ } theorem erdos\PYGZus{}1054.parts.iii :
\PYG{g+gd}{\PYGZhy{}      atTop.limsup (fun n ↦ (f n : EReal) / n = ⊤) ↔ answer(sorry) := by}
\PYG{g+gi}{+      atTop.limsup (fun n ↦ (f n : EReal) / n) = ⊤ ↔ answer(sorry) := by}
\PYG{+w}{ } sorry
\end{Verbatim}
    \caption{A \emph{Syntactic error} involving misplaced parentheses (cf.\ PR~\#1338).}
    \label{fig:misform_syntactic_erdos1054}
\end{figure}

\subsubsection{Semantic Errors}

Figures 
\ref{fig:misform_semantic_coeffs}, \ref{fig:misform_semantic_growth}, and \ref{fig:misform_semantic_covering} illustrate various semantic errors encountered during formalization.

\begin{figure}[h]
    \centering
\begin{Verbatim}[commandchars=\\\{\}]
\PYG{+w}{ } /\PYGZhy{}\PYGZhy{} `Polynomial.HasOddCoeffs f` means that all coefficients of
\PYG{+w}{ } `f : Polynomial ℤ` are odd. \PYGZhy{}/
\PYG{+w}{ } def Polynomial.HasOddCoeffs (f : Polynomial ℤ) : Prop :=
\PYG{g+gd}{\PYGZhy{}   ∀ i : ℕ, Odd (f.coeff i)}
\PYG{g+gi}{+   ∀ i ≤ f.natDegree, Odd (f.coeff i)}
\end{Verbatim}
    \caption{A \emph{Semantic error} which used an incorrect domain of quantification since the coefficients are always zero above the degree (cf.\ PRs~\#349, \#1260).}
    \label{fig:misform_semantic_coeffs}
\end{figure}

\begin{figure}[h]
    \centering
\begin{Verbatim}[commandchars=\\\{\}]
\PYG{+w}{ } /\PYGZhy{}\PYGZhy{} A group `HasPolynomialGrowth` if there exists a finite generating set
\PYG{+w}{ }   such that the growth function is bounded above by a polynomial. \PYGZhy{}/
\PYG{+w}{ } def HasPolynomialGrowth (G : Type*) [Group G] : Prop :=
\PYG{+w}{ }   ∃ (S : Set G), Set.Finite S ∧ Subgroup.closure S = ⊤ ∧
\PYG{+w}{ }     ∃ (C : ℝ) (d : ℕ), C \PYGZgt{} 0 ∧
\PYG{g+gd}{\PYGZhy{}     ∀ n : ℕ, (GrowthFunction S n : ℝ) ≤ C * (n : ℝ) \PYGZca{} d}
\PYG{g+gi}{+     ∀ n \PYGZgt{} 0, (GrowthFunction S n : ℝ) ≤ C * (n : ℝ) \PYGZca{} d}
\end{Verbatim}
    \caption{A \emph{Semantic error}; the formalization failed to account for the behavior at the natural number $0$ (cf.\ PR~\#1259).}
    \label{fig:misform_semantic_growth}
\end{figure}

\begin{figure}
    \centering
\begin{Verbatim}[commandchars=\\\{\}]
\PYG{g+gi}{+ /\PYGZhy{}\PYGZhy{} An exact covering of a group `G` is a finite collection of subgroups}
\PYG{g+gi}{+ `\PYGZob{}H\PYGZus{}1, ..., H\PYGZus{}k\PYGZcb{}` and representative `\PYGZob{}g\PYGZus{}1, ..., g\PYGZus{}k\PYGZcb{}` such that the}
\PYG{g+gi}{+ cosets `g\PYGZus{}iH\PYGZus{}i` are pairwise disjoint and their union covers `G`.}
\PYG{g+gi}{+}
\PYG{g+gi}{+ Note that this differs from `Partition (α := Subgroup G)` because the}
\PYG{g+gi}{+ covering condition there invokes `Subgroup.sup` which is subgroup generation}
\PYG{g+gi}{+ and thus stronger than union. This definition is easier to use in this}
\PYG{g+gi}{+ context than the alternative `Partition (α := Set G)`, which lacks}
\PYG{g+gi}{+ subgroup definitions such as `Subgroup.index`. \PYGZhy{}/}
\PYG{g+gi}{+ structure Group.ExactCovering (G : Type*) [Group G]}
\PYG{g+gi}{+     (ι : Type*) [Fintype ι] where}
\PYG{g+gi}{+   parts : ι → Subgroup G}
\PYG{g+gi}{+   reps : ι → G}
\PYG{g+gi}{+   nonempty (i : ι) : (parts i : Set G).Nonempty}
\PYG{g+gi}{+   disjoint : (Set.univ (α := ι)).PairwiseDisjoint}
\PYG{g+gi}{+     fun (i : ι) ↦ reps i • (parts i : Set G)}
\PYG{g+gi}{+   covers : ⋃ i, reps i • (parts i : Set G) = Set.univ}

\PYG{+w}{ } /\PYGZhy{}\PYGZhy{}
\PYG{+w}{ } If `G` is a group then can there exist an exact covering of `G` by more than
\PYG{+w}{ } one cosets of different sizes? (i.e. each element is contained in exactly
\PYG{+w}{ } one of the cosets.)
\PYG{+w}{ } \PYGZhy{}/
\PYG{+w}{ } @[category research open, AMS 20]
\PYG{g+gd}{\PYGZhy{} theorem erdos\PYGZus{}274 (G : Type*) [Group G] (hG : 1 \PYGZlt{} ENat.card G) :}
\PYG{g+gd}{\PYGZhy{}     (∃ (P : Partition (⊤ : Subgroup G)),}
\PYG{g+gd}{\PYGZhy{}       1 \PYGZlt{} P.parts.ncard ∧}
\PYG{g+gd}{\PYGZhy{}       (∀ A ∈ P.parts, ∃ᵉ (s : G) (H : Subgroup G), s • (H : Set G) = A) ∧}
\PYG{g+gd}{\PYGZhy{}       P.parts.Pairwise fun A B ↦ \PYGZsh{}A ≠ \PYGZsh{}B) ↔ answer(sorry) := by}
\PYG{g+gi}{+ theorem erdos\PYGZus{}274 : (∃ (G : Type*) (\PYGZus{} : Group G), 1 \PYGZlt{} ENat.card G ∧}
\PYG{g+gi}{+     ∃ (ι : Type*) (\PYGZus{} : Fintype ι) (P : Group.ExactCovering G ι),}
\PYG{g+gi}{+       1 \PYGZlt{} Fintype.card ι ∧ ∀ i j, i ≠ j → \PYGZsh{}(P.parts i) ≠ \PYGZsh{}(P.parts j)) ↔}
\PYG{g+gi}{+         answer(sorry) := by}
\PYG{+w}{ } sorry
\end{Verbatim}
    \caption{A \emph{Semantic error} where the formalization failed to account for the fact that \fcLeanInline{fcLean9} invokes \fcLeanInline{fcLean23} on subgroups, which gives subgroup generation rather than union of underlying sets, alongside moving \fcLeanInline{fcLean8} into an existential quantifier (cf.\ PRs~\#1262, \#5805).}
    \label{fig:misform_semantic_covering}
\end{figure}

\clearpage 

\subsubsection{Misrepresentation Errors} 

Figure \ref{fig:misform_misrep_hypothesis} provides an example of a misrepresentation error where a hypothesis from the informal text was omitted, and Figure \ref{fig:misform_misrep_quantifier} shows an error involving incorrect quantification.

\begin{figure}[h]
    \centering
\begin{Verbatim}[commandchars=\\\{\}]
\PYG{+w}{ } /\PYGZhy{}\PYGZhy{}
\PYG{+w}{ } Let \PYGZdl{}G\PYGZdl{} and \PYGZdl{}H\PYGZdl{} be finite groups of the same order with
\PYG{+w}{ } \PYGZdl{}\PYGZbs{}sum\PYGZus{}\PYGZob{}g \PYGZbs{}in G\PYGZcb{} \PYGZbs{}phi(|g|) = \PYGZbs{}sum\PYGZus{}\PYGZob{}h \PYGZbs{}in H\PYGZcb{} \PYGZbs{}phi(|h|)\PYGZdl{},
\PYG{+w}{ } where \PYGZdl{}\PYGZbs{}phi\PYGZdl{} is the Euler totient function. Suppose that \PYGZdl{}G\PYGZdl{} is simple. Is
\PYG{+w}{ } \PYGZdl{}H\PYGZdl{} necessarily simple?
\PYG{+w}{ } \PYGZhy{}/
\PYG{+w}{ } @[category research open, AMS 20]
\PYG{+w}{ } theorem kourovka.«19.25» : answer(sorry) ↔
\PYG{+w}{ }     ∀ (G H : Type) [Group G] [Group H] [Fintype G] [Fintype H],
\PYG{g+gi}{+       Fintype.card G = Fintype.card H →}
\PYG{+w}{ }        ∑ g : G, φ (orderOf g) = ∑ h : H, φ (orderOf h) →
\PYG{+w}{ }        IsSimpleGroup G → IsSimpleGroup H := by
\PYG{+w}{ }   sorry
\end{Verbatim}
    \caption{A \emph{Misrepresentation error}; the prior version was missing a hypothesis that was present in the informal text, making the statement disprovable (cf.\ PR~\#4607).}
    \label{fig:misform_misrep_hypothesis}
\end{figure}

\begin{figure}[h]
    \centering
\begin{Verbatim}[commandchars=\\\{\}]
\PYG{+w}{ } @[category research open, AMS 11]
\PYG{+w}{ } theorem erdos\PYGZus{}944 :
\PYG{g+gd}{\PYGZhy{}     (∀ k ≥ 4, ∀ r ≥ 1, ∃ (G : SimpleGraph V), G.IsErdos944 k r) ↔}
\PYG{g+gd}{\PYGZhy{}       answer(sorry) := by}
\PYG{g+gi}{+     (∀ k ≥ 4, ∀ r ≥ 1, ∃ (V : Type u) (G : SimpleGraph V), G.IsErdos944 k r) ↔}
\PYG{g+gi}{+       answer(sorry) := by}
\PYG{+w}{ } sorry
\end{Verbatim}
    \caption{A \emph{Misrepresentation error} where the vertex type \fcLeanInline{fcLean12} was incorrectly quantified, occurring as \fcLeanInline{fcLean12} was a section variable and appeared out of scope of the existential quantifier (cf.\ PR~\#1156).}
    \label{fig:misform_misrep_quantifier}
\end{figure}

\subsubsection{Implicit Conventions}

Examples of errors stemming from implicit conventions in the source material can be seen in Figures \ref{fig:misform_implicit_asymptotic} and \ref{fig:misform_implicit_induced}.

\stepcounter{footnote}%
\begin{figure}[h]
    \centering
\begin{Verbatim}[commandchars=\\\{\}]
\PYG{+w}{ } theorem green\PYGZus{}58 :
\PYG{+w}{ }   answer(sorry) ↔
\PYG{g+gd}{\PYGZhy{}      ∀ᵉ (N ≥ (1 : ℕ)) (A ⊆ Finset.Icc 1 N) (B ⊆ Finset.Icc 1 N),}
\PYG{g+gi}{+      ∀ᶠ (N : ℕ) in Filter.atTop, ∀ᵉ (A ⊆ Finset.Icc 1 N) (B ⊆ Finset.Icc 1 N),}
\PYG{+w}{ }       (N : ℝ) \PYGZca{} (0.49 : ℝ) ≤ (A.card : ℝ) →
\PYG{+w}{ }       (N : ℝ) \PYGZca{} (0.49 : ℝ) ≤ (B.card : ℝ) →
\PYG{+w}{ }       ∃ m ∈ (A + B), m.Composite := by
\PYG{+w}{ }   sorry
\end{Verbatim}
    \caption[]{An \emph{Implicit convention}; the source\footnotemark[\value{footnote}] implicitly poses the question for sufficiently large $N$ (cf.\ PR~\#3528).}
    \label{fig:misform_implicit_asymptotic}
\end{figure}
\footnotetext[\value{footnote}]{\url{https://people.maths.ox.ac.uk/greenbj/papers/open-problems.pdf}}
\newpage
\addtocounter{footnote}{2}%
\begin{figure}[h]
    \centering
\begin{Verbatim}[commandchars=\\\{\}]
\PYG{+w}{ } @[category research open, AMS 5]
\PYG{+w}{ } theorem erdos\PYGZus{}128 :
\PYG{g+gd}{\PYGZhy{}     ((∀ (G\PYGZsq{} : G.Subgraph) [Fintype G\PYGZsq{}.verts] [Fintype G\PYGZsq{}.edgeSet],}
\PYG{g+gd}{\PYGZhy{}         letI n := Fintype.card V;}
\PYG{g+gd}{\PYGZhy{}         2 * G\PYGZsq{}.verts.toFinset.card ≥ n →}
\PYG{g+gd}{\PYGZhy{}         50 * G\PYGZsq{}.edgeSet.toFinset.card \PYGZgt{} n\PYGZca{}2) → ¬ (G.CliqueFree 3))}
\PYG{g+gi}{+     (∀ (V : Type) [Fintype V] (G : SimpleGraph V),}
\PYG{g+gi}{+       (∀ V\PYGZsq{} : Set V, 2 * V\PYGZsq{}.ncard + 1 ≥ Fintype.card V →}
\PYG{g+gi}{+         50 * (G.induce V\PYGZsq{}).edgeSet.ncard \PYGZgt{} Fintype.card V \PYGZca{} 2) →}
\PYG{g+gi}{+       ¬ G.CliqueFree 3)}
\PYG{+w}{ }     ↔ answer(sorry) := by
\PYG{+w}{ } sorry
\end{Verbatim}
    \caption[]{An \emph{Implicit convention}; the source\footnotemark[\numexpr\value{footnote}-1\relax] wrote ``subgraph'' rather than ``induced subgraph'' (cf.\ PR~\#1151). The comments\footnotemark[\value{footnote}] demonstrate that the problem is trivial if one does not restrict to induced subgraphs, which is why we class this as implicit convention. It might also be considered to be a reporting error since the original paper explicitly writes ``induced subgraph'' \citep[p.~344]{Erdos93}. Note that there is also a misrepresentation error in the incorrect quantification of section variables and the cardinality inequality (cf.\ PRs~\#1390, \#4119, \#4501).}
    \label{fig:misform_implicit_induced}
\end{figure}
\footnotetext[\numexpr\value{footnote}-1\relax]{\url{https://www.erdosproblems.com/history/128}}
\footnotetext[\value{footnote}]{\url{https://www.erdosproblems.com/forum/thread/128}}
 
\subsubsection{Reporting Errors}

Figure \ref{fig:misform_reporting_erdos918} demonstrates a reporting error caused by ambiguities in the historical source material.

\stepcounter{footnote}%
\begin{figure}[H]
    \centering
\begin{Verbatim}[commandchars=\\\{\}]
\PYG{+w}{ } /\PYGZhy{}\PYGZhy{} Is there a graph with \PYGZdl{}\PYGZbs{}aleph\PYGZus{}2\PYGZdl{} vertices and chromatic number \PYGZdl{}\PYGZbs{}aleph\PYGZus{}2\PYGZdl{}
\PYG{+w}{ } such that every
\PYG{g+gd}{\PYGZhy{}  subgraph on \PYGZdl{}\PYGZbs{}aleph\PYGZus{}1\PYGZdl{} vertices has chromatic number \PYGZdl{}\PYGZbs{}aleph\PYGZus{}0\PYGZdl{}? \PYGZhy{}/}
\PYG{g+gi}{+  subgraph on \PYGZdl{}\PYGZbs{}aleph\PYGZus{}1\PYGZdl{} vertices has chromatic number \PYGZdl{}\PYGZbs{}leq\PYGZbs{}aleph\PYGZus{}0\PYGZdl{}? \PYGZhy{}/}
\PYG{+w}{ } \PYGZhy{}\PYGZhy{} Formalisation note: source material [ErHa68b] uses only induced subgraphs
\PYG{+w}{ } @[category research open, AMS 5]
\PYG{+w}{ } theorem erdos\PYGZus{}918.parts.i :
\PYG{+w}{ }     (∃ (V : Type u) (G : SimpleGraph V),
\PYG{+w}{ }       \PYGZsh{}V = ℵ\PYGZus{} 2 ∧ G.chromaticCardinal = ℵ\PYGZus{} 2 ∧
\PYG{g+gd}{\PYGZhy{}       ∀ (W : Set V) (\PYGZus{} : \PYGZsh{}W = ℵ₁), (G.induce W).chromaticCardinal = ℵ₀) ↔}
\PYG{g+gi}{+       ∀ (W : Set V) (\PYGZus{} : \PYGZsh{}W = ℵ₁), (G.induce W).chromaticCardinal ≤ ℵ₀) ↔}
\PYG{+w}{ }     answer(sorry) := by
\PYG{+w}{ }   sorry
\end{Verbatim}
    \caption[]{A \emph{Reporting error} occurring in an internal draft formalization of the problem. This problem appears in \citet{EH68} and \citet{Erdos69}. The former explicitly uses $\leq \aleph_0$, while the latter does not. This ambiguity was only clarified post draft formalization with the discovery of trivial solutions---in the (non-induced) subgraph case by the website comments\footnotemark[\value{footnote}] and by AlphaProof in the induced subgraph case. Note also that \citet{EH68} explicitly writes \emph{induced} subgraph while \citet{Erdos69} and the website do not.}
    \label{fig:misform_reporting_erdos918}
\end{figure}
\footnotetext[\value{footnote}]{\url{https://www.erdosproblems.com/forum/thread/918}}

\subsubsection{Mathematical Errors}

Figure~\ref{fig:misform_math_erdos728} illustrates a source-level mathematical error (Erd\H{o}s Problem~728), where the informal source omitted both the requirement that $\varepsilon$ be sufficiently small and the upper bound constraining $a + b - n$ to the $O(\log n)$ regime, rendering the literal statement trivial without the added upper-bound hypothesis.

\stepcounter{footnote}%
\begin{figure}[H]
    \centering
\begin{Verbatim}[commandchars=\\\{\},codes={\catcode`\$=3\catcode`\^=7\catcode`\_=8\relax}]
\PYG{+w}{ } @[category research open, AMS 11]
\PYG{+w}{ } theorem erdos\PYGZus{}728 :
\PYG{g+gd}{\PYGZhy{}     (∀ (ε C : ℝ) (hε : 0 \PYGZlt{} ε) (hC : 0 \PYGZlt{} C), ∃ a b n : ℕ,}
\PYG{g+gi}{+     (∀ᶠ ε : ℝ in }\PYG{esc}{\PYG{g+gi}{\symbol{"1D4DD}}}\PYG{g+gi}{[\PYGZgt{}] 0, ∀ C \PYGZgt{} (0 : ℝ), ∀ C\PYGZsq{} \PYGZgt{} C, ∃ a b n : ℕ,}
\PYG{+w}{ }       0 \PYGZlt{} n ∧ ε * n \PYGZlt{} a ∧ ε * n \PYGZlt{} b ∧
\PYG{+w}{ }       a ! * b ! ∣ n ! * (a + b \PYGZhy{} n)! ∧
\PYG{g+gd}{\PYGZhy{}       a + b \PYGZgt{} n + C * Real.log n) ↔ answer(sorry) := by}
\PYG{g+gi}{+       a + b \PYGZgt{} n + C * log n ∧ a + b \PYGZlt{} n + C\PYGZsq{} * log n) ↔ answer(sorry) := by}
\PYG{+w}{ }   sorry
\end{Verbatim}
    \caption[]{A \emph{Mathematical error} in the informal source statement,\footnotemark[\value{footnote}] which omitted the upper bound $a + b < n + C' \log n$ and small-$\varepsilon$ regime needed to make the problem non-trivial (cf.\ PR~\#1421).}
    \label{fig:misform_math_erdos728}
\end{figure}
\footnotetext[\value{footnote}]{\url{https://www.erdosproblems.com/forum/thread/728}}

\section{Experimental Evaluation Details}\label{sec:app_experimental_evaluation}

\subsection{Frozen Subsets}\label{sec:app_bench_lists}

The frozen subsets \texttt{FC100SolvedSet1} and \texttt{FC100OpenSet1} consist of 100 problems each, sampled uniformly at random from all statements not tagged \texttt{research open} (50 \texttt{research solved}, 34 \texttt{test}, 9 \texttt{API}, 7 \texttt{textbook}) and from \texttt{research open} statements, respectively.
Despite uniform statement-level sampling, both sets exhibit minimal clustering and broad topical coverage: \texttt{FC100OpenSet1} spans 88 distinct source files and 19 AMS subjects, while \texttt{FC100SolvedSet1} spans 81 distinct source files and 17 AMS subjects.
They are defined by the files \path{FormalConjectures/Subsets/FC100OpenSet1.lean} and \path{FormalConjectures/Subsets/FC100SolvedSet1.lean} in the repository, which import exactly the corresponding theorem statements.
These files are compiled as part of every tagged release, ensuring that the problem sets remain well-defined and their statements compile correctly across all supported Lean versions.

The subset names (e.g., \texttt{FC100OpenSet1}) are \emph{orthogonal} to the repository's release tags (\texttt{bench-v\textit{N}-lean4.\textit{X}.\textit{Y}}; see Section~\ref{sec:versioning}).
When the bench version $N$ stays the same and only the Lean version changes (e.g., from \texttt{bench-v1-lean4.27.0} to \texttt{bench-v1-lean4.29.0}), the problem set is guaranteed to contain the same statements; the new tag only ensures compilation against the updated Lean toolchain and Mathlib.
When the bench version increments (e.g., from \texttt{bench-v1} to \texttt{bench-v2}), statements in an existing set may receive fixes (e.g., corrected misformalizations) or category updates (e.g., an \texttt{open} problem reclassified as \texttt{solved}), but the \emph{membership} of the set remains stable: no problems are added or removed.
For full reproducibility, results should specify both the set name and the release tag, e.g., \texttt{FC100SolvedSet1@bench-v1-lean4.27.0}.

Building on these initial random samples, we have also released a second pair of frozen 100-problem subsets, \texttt{FC100OpenSet2} and \texttt{FC100SolvedSet2}, which use targeted curation rather than uniform random sampling.
Specifically, \texttt{FC100OpenSet2} is selected based on mathematical impact and high confidence in formalization correctness, while \texttt{FC100SolvedSet2} is curated from \texttt{research solved} statements for which no formal proof existed at release.
In the future, we plan to continue releasing additional sets (e.g., \texttt{FC500OpenSet3}) to support targeted evaluations at different scales and difficulty levels while preserving earlier sets.

\subsection{Illustrative Evaluation Details}\label{sec:illustrative_evaluation}

As noted in Section~\ref{sec:experimental_results}, our evaluations are intended to be representative rather than an exhaustive leaderboard, focusing on demonstrating the benchmark's utility and scaling signal; because optimally configuring every external prover is non-trivial, we actively encourage external baseline evaluations alongside community problem and proof contributions.
For all systems, a solution is correct if and only if the Lean 4 kernel accepts the proof term without relying on forbidden axioms (subject to the soundness of the kernel itself, e.g., a Lean Zulip thread\footnote{\url{https://leanprover.zulipchat.com/\#narrow/channel/270676-lean4/topic/Soundness.20bug.3A.20hasLooseBVars.20is.20not.20conservative/with/520153084}} discussing a now-fixed exploit).

\paragraph{Gemini 3.1 Pro Setup.}
We evaluated vanilla Gemini~3.1~Pro in a single-turn, 0-shot setting (up to 64k tokens) across ten independent runs.
For each problem, the prompt included the Lean~4 imports and theorem statement, instructing the model to output a complete, compiling Lean~4 proof.

\paragraph{AlphaProof Setup.}
We evaluated a slightly updated version of AlphaProof \citep{AlphaProof25} (3B parameters), with tree-search inference only and without the test-time reinforcement learning loop.
As reported in Table~\ref{tab:results_bench_table}, we evaluate two compute configurations across ten independent single-attempt runs per problem (reporting the mean and standard deviation across the ten attempts) to demonstrate sensitivity to inference-time search scaling: \texttt{1k sims} utilizes 1,000 simulations per problem (approximately 0.05~TPUh per attempt on v6e TPUs, or 0.5~TPUh across all ten attempts); and \texttt{16k sims} represents a $16\times$ compute increase at 16,000 simulations per problem (approximately 0.8~TPUh per attempt on v6e TPUs, or 8.0~TPUh across all ten attempts).

\paragraph{AlphaProof Nexus Setup.}
The AlphaProof Nexus \citep{DMProverAgent26} results were obtained using a development version of the method under active experimentation.
Based on current API pricing, this evaluation costs up to \$100 per problem.

\paragraph{Claude Setup.}
We evaluated Claude Sonnet 5 and Claude Opus 5 in both 0-shot and tool-use settings (four configurations total), each allowing a single attempt per problem in \texttt{FC100SolvedSet1}, with reasoning set to ``high''.
In 0-shot, results are averaged over ten independent runs during which tools are disabled and filesystem access, edits, commands, and Lean checks are prohibited.
The tool-use setting involved only a single run, allowing all of the above, with the Lean LSP\footnote{\url{https://github.com/oOo0oOo/lean-lsp-mcp}} tools available.
In all settings, proofs of the problems attempted were masked so that pre-existing solutions were not available to the agent.

\paragraph{GPT Setup.}
We evaluated GPT 5.6 Luna, Terra, and Sol on \texttt{FC100SolvedSet1}, using the Codex harness, with reasoning set to ``high''. 
Terra and Sol were evaluated in the 0-shot setting, while Luna was evaluated in both the 0-shot and tool-use settings.
In 0-shot, results are averaged over ten independent runs during which tools are disabled and filesystem access, edits, commands, and Lean checks are prohibited.
The tool-use setting involved only a single run, allowing all of the above.
In all settings, proofs of the problems attempted were masked so that pre-existing solutions were not available to the agent.

\paragraph{Leanstral Setup.}
We evaluated Leanstral 1.5 on \texttt{FC100SolvedSet1}. The model was run using the Mistral Vibe harness, with reasoning level set to ``high''. The model was allowed to use tools, including the command line, but had no access to the internet, and the proofs of problems attempted were masked.

\section{Extended Discussion}\label{sec:extended_discussion}

\subsection{Future Work}\label{sec:app_future_work}

We plan to continuously extend Formal Conjectures with new problems unblocked by newly added Mathlib definitions or from community contributions.
While our implementation uses Lean~4, the methodology is language-agnostic and portable to systems like Isabelle or Rocq should they develop comparable library support. We will also track future Lean updates to maintain ecosystem compatibility.
To better evaluate the difficulty and impact of conjectures, we plan to extend the metadata by recording when a conjecture was first proposed.
This objective metric will enable users to filter for long-standing open problems versus contemporary questions.
However, since difficulty and importance are often subjective, we also propose an interactive interface where the community can share insights.
Users will be able to vote on a conjecture's perceived difficulty, its mathematical significance, and their conviction regarding its truth value
(e.g., True, False, or independent of ZFC).

Furthermore, references to the mathematical literature are currently handled inconsistently across the repository, often requiring duplication in each file. We plan to improve this by centralizing reference management in future iterations.

This vision is being integrated into an accompanying website\footnote{\url{https://google-deepmind.github.io/formal-conjectures/}} for the repository.
The website already serves as a frontend for the benchmark, allowing users to explore the source code, view \LaTeX-rendered docstrings, and search, sort, or filter conjectures.
It also contains metadata recording when a \lstinline[language={}]{research open} problem gets solved or when a \lstinline[language={}]{research solved} problem receives a formal proof; in the future, we plan to make these changes more prominently visible.

Moving forward, we plan to expand this platform into a collaborative hub with interactive community features, including a public leaderboard to track successful AI-based solutions and recognize groups taking on these problems.

\end{document}